\documentclass[square,numbers]{article}

\usepackage[preprint]{neurips_2024}

\usepackage[utf8]{inputenc} 
\usepackage[T1]{fontenc}    
\usepackage{hyperref}       
\usepackage{url}            
\usepackage{booktabs}       
\usepackage{amsfonts}       
\usepackage{nicefrac}       
\usepackage{microtype}      
\usepackage{xcolor}         
\usepackage{times}
\usepackage{soul}
\usepackage{graphicx}
\usepackage{amsmath}
\usepackage{amsthm}
\usepackage{amssymb}
\usepackage{algorithm}
\usepackage{algorithmic}

\usepackage{multirow}
\usepackage{wrapfig}
\usepackage{subcaption}     
\usepackage{colortbl}

\title{Beyond Static Interpretability:\\Anticipating Post-SFT Mechanisms from Pre-SFT Parameters for Better Tuning}

\author{%
  Hang Chen \\
  College of Computing and Data Science\\
 Nanyang Technological University\\
  \texttt{chen.hang@ntu.edu.sg} \\
  \And
   Jiaying Zhu\\
  School of Computer Science and Engineering\\
 The Chinese University of Hong Kong\\
  \texttt{jyzhu24@cse.cuhk.edu.hk} \\
  \And
  Wenya Wang\thanks{Corresponding author}\\
  College of Computing and Data Science\\
 Nanyang Technological University\\
  \texttt{wangwy@ntu.edu.sg} \\
}

\begin{document}

\maketitle

\begin{abstract}

Mechanistic Localization bridges mechanistic interpretability and post-training optimization by isolating critical parameters via interpretative approaches and then guiding parameter-efficient Supervised Fine-Tuning (SFT) in a ``locating-then-tuning'' paradigm. However, due to the retrospective nature of mechanistic interpretability, directly interpreting pre-SFT models introduces misleading conclusions. Specifically for novel tasks, initially identified neurons differ drastically from those governing the final model, introducing biases that actively disrupt SFT. To address this, we propose a forward-looking localization framework that accurately estimates the post-SFT interpretability state using only pre-SFT parameters and the target dataset. Theoretically, we model SFT as a continuous parameter evolution, leveraging Taylor expansion to rigorously bridge the post-tuning mechanistic objective with the pre-SFT model's dynamic gradients. Practically, we design dual-granularity (neuron- and component-level) localization pipelines. Extensive experiments demonstrate that our approach not only provides superior SFT guidance but also exhibits robust performance and temporal scalability across increasing model sizes. This work transcends the fundamental limitation of traditional interpretability—its inability to identify task-critical mechanisms before they are trained—pioneering a predictive frontier that unites mechanistic interpretability with targeted optimization. Code is available at: \url{https://github.com/Zodiark-ch/Future_localization}. 

\end{abstract}
\section{Introduction}

As a pivotal foundation for reverse-engineering LLMs, mechanistic interpretability illuminates internal execution mechanisms by evaluating the causal effects of specific components or neurons through counterfactual analysis~\citep{heimersheim2024use,rai2024practical,sharkey2025open,somvanshi2026bridging}. Beyond passive diagnostic insights, its rigorous causal framework offers a principled basis for guiding post-training optimization. In particular, the \emph{locating-then-tuning} paradigm~\citep{chen2026navigatingoldmapspitfalls} successfully operationalizes this insight into SFT-based adaptation: it first employs \emph{mechanistic localization}~\citep{guo2025mechanistic} on the \textbf{pre-SFT} base model to identify key parameters governing the target task, and subsequently restricts parameter-efficient tuning only to these identified key parameters to yield the final \textbf{post-SFT} model. By preventing unconstrained weight updates from corrupting pre-existing representations, this paradigm has exhibited immense promise in critical domains including LLM unlearning~\citep{jia2024wagle,chen2026clue}, knowledge editing~\citep{pmlr-v202-panigrahi23a,zhang2024comprehensive}, and transparent post-training~\citep{wang2025towards,yin2024lofit,jing2026guidingllmposttrainingdata,zhang2026locate}.

However, relying on the pre-SFT model for mechanistic localization introduces a fundamental flaw: because the model has not yet been adapted to the target task, the parameters identified as task-relevant in pre-SFT model may not reflect the mechanisms that emerge during SFT, potentially leading to misleading localization for subsequent fine-tuning~\citep{chen2026navigatingoldmapspitfalls}. Consider an extreme yet representative case where the target task is entirely novel to the pre-SFT model. Because the pre-SFT model inherently lacks the mechanistics required to execute this unseen task, mechanistic interpretability on the pre-SFT model fails to identify meaningful task-governing parameters. Consequently, it yields deceptive signals—inadvertently freezing parameters vital for acquiring new skills and obstructing necessary circuit evolution~\citep{prakash2024finetuning,jain2024mechanistically,chhabra-etal-2025-neuroplasticity}. Fundamentally, this limitation stems from an intrinsic tension between the prospective requirement of tuning and the retrospective nature of interpretability: locating-then-tuning demands that interpretability \emph{anticipate} which parameters will be recruited to learn the target skill, yet mechanistic interpretability can only dissect capabilities that are \emph{already instantiated} within existing parameters. This creates a classic ``chicken-and-egg'' dilemma—one must localize the task-critical parameters before tuning, yet the corresponding circuits only emerge after the model has learned the task.

To this end, this paper aims to challenge the retrospective practice of interpretability area by answering a fundamental question: \emph{how can we predict the neurons (or components) ($\mathcal{N}(\theta^{I})$) responsible for a target task in an ideal post-SFT model($\theta^{I}$) using only the pre-SFT model parameters ($\theta$)?} 

We view fine-tuning ($\theta \rightarrow \theta^{I}$) as an optimization establishing an initial gradient direction and subsequently traversing a long distance toward a local optimum. Consequently, we decompose $\theta \rightarrow \mathcal{N}(\theta^{I})$ into \textbf{two theoretical steps}: isolating ``\textbf{direction}'' and extrapolating ``\textbf{distance}''. First, we formulate $\theta \rightarrow \mathcal{N}(\theta')$, where $\theta'$ represents the parameters following an infinitesimal update toward the target task's supervision. This decouples the update distance to exclusively capture the initial direction. Building upon $\mathcal{N}(\theta')$, the second step achieves $\mathcal{N}(\theta') \rightarrow \mathcal{N}(\theta^{I})$ by conceptualizing the total SFT parameter shift, $\Delta \theta = \theta^{I} - \theta$, as the accumulation of infinite recursive updates originating from $\theta'$.

Our approach only requires fine-tuning on $1\%$ of the training dataset for a single epoch to derive $\theta'$. Inheriting the computational paradigm of Attribution Patching~\citep{nanda2023attribution,syed-etal-2024-attribution}, our method necessitates only two forward passes and one backward pass to collect the activation and gradient data required for the estimation. This computational overhead is entirely negligible compared to the full fine-tuning process or other interpreting approaches. Furthermore, to accommodate larger models and diverse practical scenarios, we designed two distinct estimation pipelines: a fine-grained estimation at the neuron level and a rapid estimation at the component level.

We empirically validate our approach on the Mistral-7B model, demonstrating its superior capability in guiding SFT compared to existing localization baselines. Furthermore, through rigorous ablation studies on arithmetic tasks of varying complexities, we establish the robustness of our method against increasing task difficulty. Additionally, we confirm the scalability of our approach across larger architectures, specifically LlaMA-2-13B and Qwen3-30B. Finally, we provide intrinsic validation of our method's efficacy directly through the lens of the resulting interpretability analyses.

In summary, our main contributions are twofold:
\begin{itemize}
    \item We propose a theoretical framework that leverages current parameters to estimate post-SFT mechanisms. To the best of our knowledge, this is the first work to predict future interpretability states. Confronting this question marks a transformative transition beyond traditional retrospective interpretability, pioneering a new frontier for its actionable utility in post-training optimization.
    \item We develop an estimation pipeline that achieves state-of-the-art localization performance. Validated across diverse LLM families, our method demonstrates robust scalability while maintaining highly stable computational efficiency as model parameters increase.
\end{itemize}

\section{Preliminaries}
\subsection{Task Definition}
Let $\mathcal{M}$ parameterized by $\theta$ denote the current model prior to fine-tuning, and $\mathcal{M}^{I}$ parameterized by $\theta^{I}$ represent the ideal post-tuning model. For a target task $\mathcal{T}$, let $\mathcal{D}_{\mathcal{T}}$ be its associated dataset. By definition, the current model $\mathcal{M}$ performs poorly on $\mathcal{D}_{\mathcal{T}}$, whereas the ideal model $\mathcal{M}^{I}$ achieves optimal performance on $\mathcal{D}_{\mathcal{T}}$ without degrading any pre-existing capabilities\footnote{In practice, SFT can only continuously approximate this ideal model rather than achieving it perfectly.}.

We denote the subset of parameters primarily responsible for task $\mathcal{T}$ as $\mathcal{N}^{\theta} \subset \theta$, which is typically identified via mechanistic interpretability methods. The existing \emph{locating-then-tuning} paradigm operates through the following pipeline with partial tuning only operating on $\mathcal{N}^{\theta}$:
\begin{equation*}
    \mathcal{M}(\theta), \mathcal{D}_{\mathcal{T}} \xrightarrow{\text{localization}} \mathcal{N}^{\theta} \xrightarrow[\mathcal{D}_{\mathcal{T}} ]{\text{partial tuning}} \hat{\mathcal{M}}^{I}(\theta^{I})
\end{equation*}

In this work, our objective is to reformulate the localization step to directly predict the critical parameter subset ($\mathcal{N}^{\theta^{I}}$) of the ideal model:
\begin{equation*}
    f: \langle \mathcal{M}(\theta), \mathcal{D}_{\mathcal{T}} \rangle \longrightarrow \mathcal{N}^{\theta^{I}}
\end{equation*}
\subsection{Related Work}
Existing research on the \emph{locating-then-tuning} paradigm primarily focuses on optimizing mechanistic localization, relying predominantly on gradients~\citep{jia2024wagle,yin2024lofit,pmlr-v202-panigrahi23a} or causal effects~\citep{wang2025towards,jing2026guidingllmposttrainingdata,guo2025mechanistic,chen2026clue}. Recent advances in mechanistic interpretability~\citep{heimersheim2024use,lieberum2023does,chen2026rethinking,chen2026skill} demonstrate that, under rigorous experimental setups, causal effect-based approaches provide a more accurate and comprehensive identification of task-critical neurons. Specifically, these methods quantify the causal effect for the activation of a target parameter, neuron, or component $a$ under the target dataset $\mathcal{D}_{\mathcal{T}}$,  by measuring the output discrepancy when the activation is replaced with a counterfactual value. This is formalized as:
\begin{equation*}
    \Delta E_{\mathcal{D}_{\mathcal{T}}}(a) = f_a(A(x_c)) - f_a(A(x_r))
\end{equation*}
where $f_a(\cdot)$ represents an output-dependent metric (e.g., logits or loss) as the samples from $\mathcal{D}_{\mathcal{T}}$ are input, while $A(x_c)$ and $A(x_r)$ denote the clean and corrupted (counterfactual) activations, respectively. A larger $\Delta E$ signifies a stronger causal relationship with the task output. Currently, circuit discovery~\citep{conmy2023towards,bhaskar2024finding} and path patching~\citep{wanginterpretability} are the most prevalent causal methodologies employed in mechanistic localization.

Highly pertinent to our study is \emph{Attribution Patching}~\citep{nanda2023attribution,syed-etal-2024-attribution}, a causal effect method that approximates this influence using a first-order Taylor expansion:
\begin{equation*}
    \Delta E_{\mathcal{D}_{\mathcal{T}}}(a) = f_a(A(x_c)) - f_a(A(x_r)) \approx (A(x_c) - A(x_r)) \cdot \frac{\partial f_a}{\partial a}\bigg|_{a=A(x_r)}
\end{equation*}
This mathematical reformulation elegantly circumvents the computationally prohibitive exhaustive search required by standard counterfactual ablation. Instead, it extracts the necessary metrics for all parameters simultaneously requiring only two forward passes to obtain $A(x_c)$ and $A(x_r)$, and a single backward pass to compute the gradients $\frac{\partial f_a}{\partial a}\big|_{a=A(x_r)}$.

Beyond static analysis, our work is motivated by recent dynamic interpretability studies analyzing mechanistic shifts during SFT~\citep{prakash2024finetuning,jain2024mechanistically}. These works highlight that target task mechanisms undergo drastic transformations—including shifts in circuit strength~ \citep{chhabra-etal-2025-neuroplasticity}, topology~ \citep{wang2025towards}, and key nodes~\citep{chen2026navigatingoldmapspitfalls}—during tuning. As a result, deriving the critical neuron set $\mathcal{N}^{\theta}$ purely from the pre-SFT parameters $\theta$ yields a biased localization. Instead of aiding the tuning process, this static guidance misdirects optimization and deteriorates task learning, thereby motivating our framework to estimate the ideal post-SFT state.

\section{Method}

The inherent challenge in our objective lies in the pragmatic nature of mechanistic interpretability: it is traditionally a retrospective analysis conducted on the explicit parameters of already-trained models. Existing theories do not support predicting interpretability states prior to tuning. Whereas targeted SFT aims at updating parameters supposedly influential for the target task. This presents a critical challenge: \textit{given only the current parameters $\theta$ and the target dataset $\mathcal{D}_{\mathcal{T}}$, how can we estimate the importance of future model components without actually executing the entire SFT process?} 

To address this, we conceptualize SFT as a continuous optimization trajectory—starting from an initial well-trained parameter distribution, evolving along a specific gradient direction, and traversing a certain distance to reach a local optimum. By bridging interpretability with gradient dynamics, our framework solves this challenge in two intuitive steps: first determining the initial evolutionary direction, and then extrapolating it to an appropriate distance.
Specifically, first, in Section~\ref{sec3.1single}, we capture the initial gradient direction by modeling an infinitesimal update of $\theta$ towards the SFT objective on $\mathcal{D}_{\mathcal{T}}$, denoted as $\theta'$, and then we formulate its causal effect entirely using the initial $\theta$. Second, in Section~\ref{sec3.2multi}, we extrapolate this directional estimate to the optimal distance. By iteratively accumulating an infinite series of these infinitesimal updates, we approximate the causal effect of the ideal post-SFT model, $\theta^{I}$, deriving our final predictive expression.


\subsection{Single Step: From $\theta$ to $\theta '$ for Gradient Direction}\label{sec3.1single}

Let $\theta'$ denote the parameters following an infinitesimal update of $\theta$ directed by the SFT objective on $\mathcal{D}_{\mathcal{T}}$. For any arbitrary parameter $a$ under $\theta'$, its corresponding Attribution Patching formula is given by:
\begin{equation}
    f^{\theta'}_a(A(x_c)) - f^{\theta'}_a(A(x_r)) \approx (A^{\theta'}(x_c) - A^{\theta'}(x_r)) \cdot \frac{\partial f^{\theta'}_a}{\partial a^{\theta'}}\bigg|_{a=A^{\theta'}(x_r)}
\end{equation}

Evaluating the terms $A^{\theta'}(x_r)$, $A^{\theta'}(x_c)$, and the gradient $\frac{\partial f^{\theta'}_a}{\partial a^{\theta'}}\big|_{a=A^{\theta'}(x_r)}$ inherently requires two actual forward passes and one backward pass through the hypothetical $\theta'$ model, which is practically intractable. Consequently, we seek to approximate these terms by applying a first-order Taylor expansion around $\theta$. This yields a reformulation of the causal effect for $\theta'$ expressed entirely in terms of $\theta$:
\begin{equation}
    f^{\theta'}_a(A(x_c)) - f^{\theta'}_a(A(x_r)) \approx f^{\theta}_a(A(x_c)) - f^{\theta}_a(A(x_r)) + \Delta \theta \cdot S(\theta)
    \label{eqtsingle}
\end{equation}
where
\begin{equation}
    S(\theta) = \left[ \left( \frac{\partial A^{\theta}(x_c)}{\partial \theta} - \frac{\partial A^{\theta}(x_r)}{\partial \theta} \right) \cdot \frac{\partial f^{\theta}_a}{\partial a^{\theta}} + \left(A^{\theta}(x_c) - A^{\theta}(x_r)\right) \cdot \frac{\partial^2 f^{\theta}_a}{\partial a \partial \theta} \right]
\end{equation}

Here, $\Delta \theta = \theta' - \theta$ represents the distance of the parameter update. Noticeably, $S(\theta) \equiv \nabla_{\theta} \left[ (A^\theta(x_c) - A^\theta(x_r)) \cdot \frac{\partial f^{\theta}_a}{\partial a^{\theta}} \right]$. Therefore, it represents the gradient of the attribution score with respect to $\theta$, capturing the sensitivity of the target neuron $a$'s attribution value to parameter updates.

Through Equation~\ref{eqtsingle} (the detailed derivation is provided in Appendix~\ref{suppderivationeqt2}), we transform the estimation of $\theta'$'s causal effect---which previously relied on empirical forward and backward passes---into an estimation of the parameter update distance. In the following sections, we will elaborate on how to determine $\Delta \theta$ in order to ultimately approximate the interpretability state of the final ideal model, $\theta^{I}$.

\subsection{Multiple Steps: From $\theta '$ to $\theta^{I}$ for Gradient Distance}\label{sec3.2multi}
We conceptualize the transition from the initial parameters $\theta$ to the ideal parameters $\theta^I$ as an infinite sequence of continuous, additive updates. For notational clarity, we represent this evolutionary trajectory as a series of states $\theta^0, \theta^1, \dots, \theta^I$, where the initial state is defined as $\theta^0 \equiv \theta$. Within this continuum, any intermediate state $\theta^k$ effectively functions as the infinitesimal update, previously denoted as $\theta'$, relative to its immediate predecessor $\theta^{k-1}$.

Following Equation~\ref{eqtsingle}, the causal effect $\Delta E^{\theta^{k+1}}_{\mathcal{D}_{\mathcal{T}}}(a)$ at any subsequent state $\theta^{k+1}$ can be recursively formulated as:
\begin{equation}
    \Delta E^{\theta^{k+1}}_{\mathcal{D}_{\mathcal{T}}}(a) \approx \Delta E^{\theta^{k}}_{\mathcal{D}_{\mathcal{T}}}(a) + \Delta \theta^k \cdot S(\theta^k)
\end{equation}

Consequently, by accumulating these incremental updates, the causal effect for the final ideal model is given by:
\begin{equation}
    \Delta E^{\theta^{I}}_{\mathcal{D}_{\mathcal{T}}}(a) \approx \Delta E^{\theta}_{\mathcal{D}_{\mathcal{T}}}(a) + \sum_{k=0}^{I-1} \Delta \theta^k \cdot S(\theta^k)
\end{equation}
The summation term in this formulation effectively constitutes a discrete path integral along the parameter update trajectory $\gamma$. In the continuous limit, this can be formally expressed as$\int_{\gamma} S(\theta) \cdot d\theta$. 
To render this computation tractable without evaluating intermediate states, we approximate this continuous integral using the trapezoidal rule\footnote{The underlying assumption of this trapezoidal approximation is mild: as the gradient of the attribution score, $S(\theta)$ typically exhibits local monotonicity during tuning. Consequently, the introduced systematic and unidirectional error across parameters barely affects the relative ranking of each neuron's final causal effect.}, yielding:
\begin{equation}
    \Delta E^{\theta^{I}}_{\mathcal{D}_{\mathcal{T}}}(a) \approx \Delta E^{\theta}_{\mathcal{D}_{\mathcal{T}}}(a) + \frac{1}{2} \left[ S(\theta^0) + S(\hat{\theta}^I) \right] \cdot \sum_{k=0}^{I-1} \Delta \theta^k
\end{equation}


Given that $S(\theta)$ is strictly equivalent to the gradient of the attribution score with respect to $\theta$ (i.e., $S(\theta) \equiv \nabla_{\theta} [\Delta E^{\theta}_{\mathcal{D}_{\mathcal{T}}}(a)]$). By defining a differential operator $D = \Delta \theta \cdot \nabla_{\theta}$, we demonstrate that the infinite series of higher-order derivatives (Hessians, tensors) governing $S(\theta^k)$ can be rigorously folded into an exponential translation operator, $e^{k \Delta \theta \cdot \nabla_{\theta}}$. Under this functional analysis perspective, evaluating $S$ at state $k$ is mathematically equivalent to simply shifting the input parameters. Consequently, we establish that $S(\hat{\theta}^I) = S(\theta + K \Delta \theta)$, allowing us to compute the final sensitivity by merely scaling the update distance. We refer readers to Appendix~\ref{appendix:translation} for the complete theoretical proof.

Ultimately, we simplify the estimation of the causal effect for the ideal model, $\theta^I$, to the following formulation:
\begin{equation}
    \Delta E^{\theta^{I}}_{\mathcal{D}_{\mathcal{T}}}(a) \approx \Delta E^{\theta}_{\mathcal{D}_{\mathcal{T}}}(a) + \frac{1}{2} \left[ S(\theta^0) + S(\theta^0+K \cdot \Delta \theta) \right] \cdot K \cdot \Delta \theta
    \label{eqtmultistep}
\end{equation}

where $K$ denotes a step scalar that extrapolates the transition $\theta \rightarrow \theta'$ to the ideal macroscopic transition $\theta \rightarrow \theta^I$. While theoretically derived from an infinitely small step, practically, we approximate $\theta'$ using a sufficiently small parameter update. 
Moreover, the ideal post-SFT state is non-unique—varying based on optimization trajectories or random initializations—we do not treat $K$ as a single determined value. Instead, we use $K$ as an extrapolation scalar to project along the known gradient direction, sampling a range of $K$ values to explore plausible future configurations ${S(\hat{\theta}^I)}=S(\theta^0+K \cdot \Delta \theta)$ and ultimately estimating the mean causal effect of the target task mechanisms.
Therefore, this approximation requires that $\Delta\theta=\theta'-\theta$ robustly captures the general optimization direction of the target task, while maintaining $\Delta\theta$ small enough to provide the step scalar $K$ with ample degrees of freedom to explore the parameter space.


\begin{figure*}[htbp]
    \centering
    \includegraphics[width=0.9\textwidth]{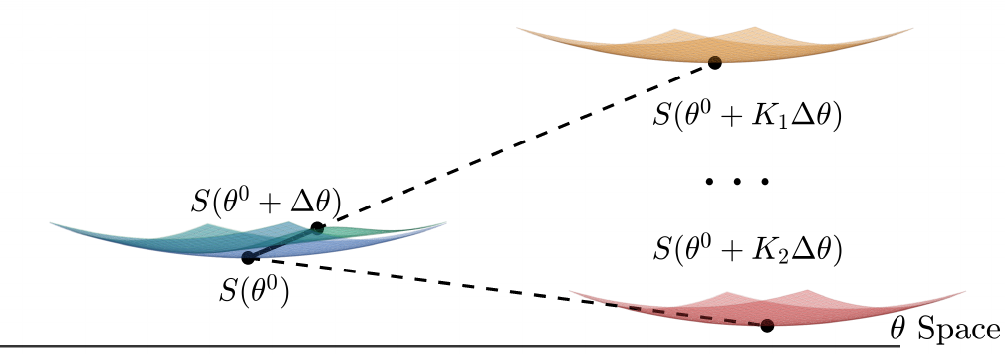}
    \caption{\textbf{Extrapolation of the ideal state via the step scalar $K$.} Different values of $K$ yield distinct sensitivity estimations $S(\theta + K \cdot \Delta \theta)$, reflecting that the ideal parameter state $\theta^I$ consists of an infinite set of valid configurations rather than a single deterministic solution.}
    \label{fig:Kmapping}
\end{figure*}

\section{Implemetation}
\subsection{Probing SFT}\label{secprobingsft}
In this work, we acquire $\theta'$ through a \emph{probing SFT} strategy. Specifically, we uniformly sample $1\%$ of the training dataset to perform a tentative fine-tuning for a single epoch at $1\%$ of the original learning rate, designating the resulting updated weights as $\theta'$. As explored in our empirical analysis (Section~\ref{secexprobing}), a larger divergence between $\theta'$ and $\theta$ does not monotonously correlate with improved performance; in fact, deriving $\theta'$ from $1\%$ of the training data yields significantly superior results compared to using $10\%$. This phenomenon occurs because $\Delta \theta$ must remain sufficiently infinitesimal to preserve adequate degrees of freedom for the extrapolated state $\theta + K \cdot \Delta \theta$, while simultaneously mitigating the influence of gradient-irrelevant noise inherent in $\theta'$. 

Notably, performing attribution patching directly on $\theta'$ (hereafter referred to as the \emph{probing model}) constitutes a compelling baseline, which we denote as \emph{Localization from $\theta'$}. It intuitively offers more foresight than localization from the initial $\theta$, yet avoids the complexity of estimating the ideal state $\hat{\theta}^I$. Mathematically, however, $\theta'$ remains proximate to $\theta$ and distant from $\hat{\theta}^I$. As formally proven in Appendix~\ref{suppproofprobing}, establishing the equivalence $\Delta E^{\hat{\theta}^{I}}_{\mathcal{D}_{\mathcal{T}}}(a) \approx \Delta E^{\theta'}_{\mathcal{D}_{\mathcal{T}}}(a)$ necessitates imposing stringently unrealistic assumptions on the SFT parameter update trajectory. That is, for any step $k$, it satisfies: $\frac{\partial (A(x_c) - A(x_r))}{\partial \theta} \cdot \Delta \theta \approx X_k - X_0$.
Nevertheless, regarding the final estimation, $\Delta E^{\theta'}_{\mathcal{D}_{\mathcal{T}}}(a)$ inherently introduces significantly less approximation error, at least from the perspective of parameter variation. Our empirical evaluation in Section~\ref{secexablation} thoroughly compares these two paradigms---\emph{Localization from $\hat{\theta}^I$} versus \emph{Localization from $\theta'$}. The results demonstrate that as the difficulty of the target SFT task increases, the superiority of \emph{Localization from $\hat{\theta}^I$} over \emph{Localization from $\theta'$} becomes increasingly pronounced.

\subsection{Locating-then-tuning Pipelines}
We present a generalized pipeline for the \emph{locating-then-tuning} paradigm. During the localization phase, we compute the causal effect $\Delta E^{\theta^{I}}_{\mathcal{D}_{\mathcal{T}}}(a)$ for all parameters and rank them in descending order to determine their relative importance. To ensure a robust estimation, we randomly sample 10 distinct values of $K$ and average their resulting scores. The optimal sampling interval for $K$ is inherently task-dependent, a phenomenon we thoroughly analyze in Appendix~\ref{sec:experiments_k_interval}, and it  explores a probing model approach to recommend $K$, and provides an analysis of the robustness concerning $K$ value sampling. Furthermore, to circumvent the prohibitive computational overhead induced by the second-order partial derivatives in $S(\theta)$, we employ a highly time-efficient approximation strategy (We show the acceptable error in Figure~\ref{fig:circuiterror}):
\begin{equation}
    \frac{\partial^2 f_a}{\partial a \partial \theta} \cdot \Delta \theta \approx \frac{\partial f^{\theta'}_a}{\partial a^{\theta'}} - \frac{\partial f^{\theta}_a}{\partial a^{\theta}}
    \label{eqt2order}
\end{equation}
During the tuning phase, we implement differential optimization strategies tailored to the derived parameter importance. Specifically, we instantiate this pipeline at two distinct granularities:

\noindent \textbf{(1) Neuron-level:} The minimal unit for computing $\Delta E^{\theta^{I}}_{\mathcal{D}_{\mathcal{T}}}(a)$ is defined as an individual neuron (i.e., a single scalar value within a parameter matrix). We selectively update only the top 20\% most critical neurons while freezing the remaining 80\%, optimizing via the standard next-token cross-entropy loss for SFT. For example, when applied to the Mistral-7B model, this strategy restricts parameter updates to exactly 1.4B out of the 7B total neurons.

\noindent \textbf{(2) Component-level:} The minimal unit is designated as an entire parameter matrix (specifically, $W_q, W_k, W_v, W_o$ within each attention head, and $W_{up}, W_{gate}, W_{down}$ in the MLP). We integrate Low-Rank Adaptation (LoRA)~\citep{hu2022lora} into this pipeline by dynamically allocating ranks according to the component importance rankings. Higher-ranked components are assigned proportionally higher LoRA ranks, scaling within a discrete range of $[1, 32]$ while maintaining an overall average rank of $8$.
\section{Experiments}
\subsection{Setups}
We employ Mistral-7B~\citep{jiang2023mistral7b} as our primary evaluation model, while further incorporating LLaMA-2-13B~\citep{touvron2023llama2openfoundation} and Qwen3-30B~\citep{yang2025qwen3technicalreport} to validate the scalability of our approach.  
We evaluate LLM fine-tuning performance across three primary domains: \textbf{Natural Language Understanding (NLU)} across the GLUE benchmark (SST-2, MRPC, QQP, MNLI, and RTE)~\citep{wang2019glue}, \textbf{Logical Reasoning (LR)} via BOOL~\citep{suzgun2023challenging}, and \textbf{Mathematical Reasoning (MR)} using Arithmetic~\citep{srivastava2023beyond}\footnote{Unlike mainstream instruction-following tasks in post-training SFT, the locating-then-tuning paradigm typically employs tasks with highly controlled output lengths (e.g., single-token generation). This constraint is intentional: it yields cleaner causal mechanisms and isolates the exact effects of mechanistic localization, allowing us to explicitly ablate the contributions of mechanistic localization. We discuss this limitation further in Section~\ref{sec:conclusion}.}. To examine localization efficacy under varying adaptation complexities, we stratify the Arithmetic benchmark into hierarchical subtasks ranging from 2-digit to 7-digit operations and 1-step to 6-step computations.
Additionally, to facilitate a granular examination of interpretability outcomes---specifically through circuit graph and component statistics analyses---we incorporate established benchmarks from the mechanistic interpretability literature: Indirect Object Identification (IOI)~\citep{wanginterpretability}, Induction~\citep{conmy2023towards}, WinoGrande~\citep{ai2:winogrande}, Genderr~\citep{mathwin2023identifying}, and Docstring~\citep{heimersheim2023circuit}. Detailed prompt configurations for these datasets are provided in Appendix~\ref{suppdetails}.

We evaluate our proposed localization methods---denoted as Ours (N) for the neuron-level pipeline and Ours (C) for the component-level pipeline---against a suite of state-of-the-art baselines. These comprise gradient-guided approaches (Graft~\citep{pmlr-v202-panigrahi23a} and WAGLE~\citep{jia2024wagle}) alongside causal effect-guided methods (CLUE~\citep{chen2026clue}, FLU~\citep{guo2025mechanistic}, and CircuitLoRA~\citep{wang2025towards}). To comprehensively assess post-tuning efficacy, we monitor two primary metrics: Target Task Accuracy (TTA) and Pervasiveness Task Accuracy (PTA). TTA is directly derived from the label accuracy on the target tasks (e.g., average accuracy from all tasks in GLUE, BOOL, and Arithmetic). Conversely, PTA quantifies the degradation of pre-existing general capabilities; it is computed as the macro-average across a diverse array of generalized benchmarks using the \texttt{lm-evaluation-harness} framework~\citep{eval-harness}.

For the fine-tuning optimization, we minimize the cross-entropy loss over the target task labels. The training hyperparameters are standardized to a learning rate of $1 \times 10^{-5}$, $3$ training epochs, and a default LoRA rank of $8$. All experiments are executed on NVIDIA RTX A6000 with corresponding CUDA environments.

\subsection{Main Results}
\begin{table}[htbp]
    \centering
    \resizebox{0.9\textwidth}{!}{
    \begin{tabular}{lcccccc}
        \toprule
        \multirow{2}{*}{\textbf{Method}} & \multicolumn{2}{c}{\textbf{LR}} & \multicolumn{2}{c}{\textbf{NLU}} & \multicolumn{2}{c}{\textbf{MR}} \\
        \cmidrule(lr){2-3} \cmidrule(lr){4-5} \cmidrule(lr){6-7}
        & \textbf{TTA} & \textbf{PTA} & \textbf{TTA} & \textbf{PTA} & \textbf{TTA} & \textbf{PTA}  \\
        \midrule
        Graft       & 82.76$_{\pm 0.42}$ & 33.27$_{\pm 0.04}$ & 84.96$_{\pm 0.08}$ & 31.24$_{\pm 0.14}$ & 24.19$_{\pm 2.13}$ & 44.31$_{\pm 0.11}$ \\
        FLU         & 88.57$_{\pm 0.35}$ & 42.57$_{\pm 0.09}$ & 84.27$_{\pm 0.02}$ & 36.72$_{\pm 0.13}$ & 87.16$_{\pm 1.07}$ & 49.52$_{\pm 0.01}$ \\
        WAGLE       & 89.55$_{\pm 0.28}$ & 44.31$_{\pm 0.06}$ & 85.28$_{\pm 0.11}$ & 43.57$_{\pm 0.04}$ & 66.39$_{\pm 1.15}$ & 51.23$_{\pm 0.09}$ \\
        CLUE        & 90.49$_{\pm 0.44}$ & 45.42$_{\pm 0.02}$ & 85.17$_{\pm 0.07}$ & 45.61$_{\pm 0.10}$ & 52.78$_{\pm 1.55}$ & 53.21$_{\pm 0.12}$ \\
        CircuitLoRA & 97.67$_{\pm 0.13}$ & 60.62$_{\pm 0.11}$ & 85.09$_{\pm 0.06}$ & 61.55$_{\pm 0.09}$ & 98.60$_{\pm 0.12}$ & 59.66$_{\pm 0.14}$ \\
        \midrule
        \textbf{Ours (N)} & 98.51$_{\pm 0.09}$ & 60.53$_{\pm 0.05}$ & 86.31$_{\pm 0.13}$ & 62.02$_{\pm 0.07}$ & 98.79$_{\pm 0.14}$ & 60.13$_{\pm 0.10}$ \\
        \textbf{Ours (C)} & \textbf{100.00}$_{\pm 0.00}$ & \textbf{62.64}$_{\pm 0.08}$ & \textbf{87.55}$_{\pm 0.04}$ & \textbf{63.95}$_{\pm 0.11}$ & \textbf{100.00}$_{\pm 0.00}$ & \textbf{61.48}$_{\pm 0.06}$ \\
        \bottomrule
    \end{tabular}
    }
    \caption{Performance comparison of Locating-then-tuning methods.}
    \label{tab:tta_pta_results}
\end{table}

Table~\ref{tab:tta_pta_results} reports the performance of Mistral-7B model of various localization methods across the LR, NLU, MR domains, where the Arithmetic results represent the average performance across the 2- to 5-digit subtasks. As demonstrated, our proposed methods establish a substantial margin over all baseline approaches in both Target Task Accuracy (TTA) and the retention of general capabilities (PTA). This confirms that estimating future parameter importance provides a distinct advantage over conventional mechanistic localization derived solely from current parameters. Interestingly, our component-level pipeline (Ours (C)) significantly outperforms its neuron-level counterpart (Ours (N)), suggesting that an increasingly fine granularity in mechanistic localization does not strictly correlate with improved performance. Furthermore, CircuitLoRA---another baseline utilizing LoRA for fine-tuning---emerges as the second-best method. This observation implies that LoRA's robust adaptability to single, formalized tasks may largely account for the empirical superiority of the component-level approach over the neuron-level configuration.

\begin{figure*}[htbp]
    \centering
    \includegraphics[width=0.8\textwidth]{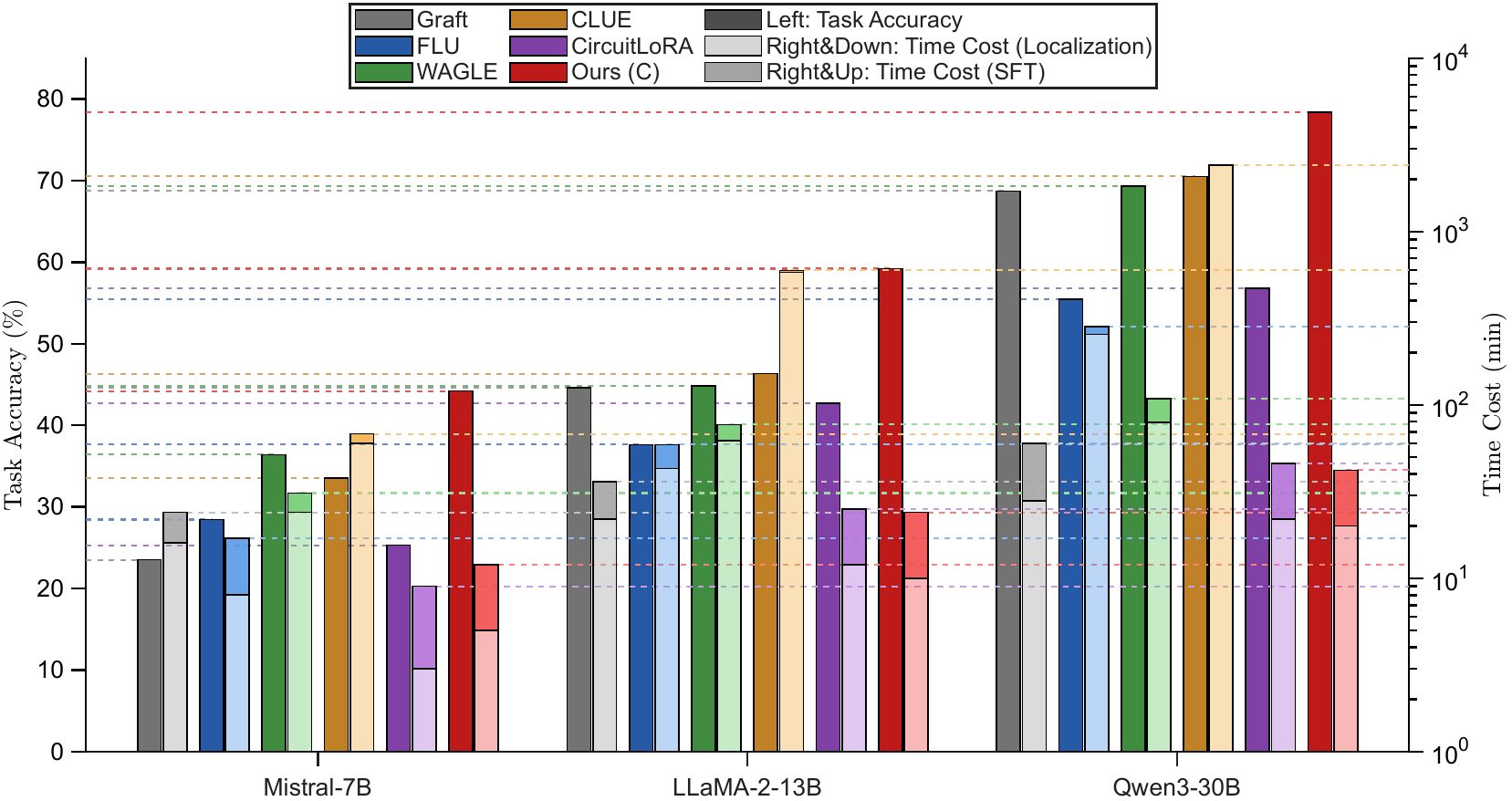}
    \caption{The scalability of performance and time cost with mode size increasing.}
    \label{fig:scalability}
\end{figure*}

Furthermore, to validate the scalability of our proposed methods, we conduct a comparative analysis across three text-to-text models of varying parameter sizes that lack native complex reasoning capabilities: Mistral-7B, LLaMA-2-13B, and Qwen3-30B. We select the 5-step and 6-step Arithmetic subtasks for evaluation, employing the mean Target Task Accuracy (TTA) as the primary performance metric. To quantify computational efficiency, we report the total time expenditure, defined as the sum of the localization and fine-tuning durations.

As illustrated in Figure~\ref{fig:scalability}, our method consistently sustains optimal fine-tuning performance as the model scale increases, demonstrating its robust efficacy on LLMs. Concurrently, the fine-tuning time cost exhibits a relatively gradual growth rate, though we acknowledge this may be partially attributed to the inherent simplicity of the chosen target tasks. Conversely, the localization time cost escalates prominently with model size. This sharp increase confirms that the computational complexity of mechanistic localization strictly exceeds $\mathcal{O}(n)$, a consequence of the auxiliary spatial and computational overhead mandated by interpretability procedures. Despite this inherent overhead, our approach maintains nearly the lowest overall computational complexity among the evaluated baselines, underscoring its practical viability for large-scale LLMs.

Finally, to further explore the broader potential of our approach, we conduct multi-task joint training experiments, as detailed in Appendix~\ref{suppmultitask}. Existing literature demonstrates that joint training frequently exacerbates inter-task conflicts~\citep{chen2026clue,yang2026localperturbationtheorycrossdomain}, a phenomenon primarily manifesting within shared neurons that are simultaneously responsible for multiple distinct tasks~\citep{elhage2022toymodelssuperposition}. By jointly evaluating task accuracy and the prevalence of these conflicting neurons, we empirically demonstrate that our proposed localization method provides crucial foresight. This predictive capability substantially mitigates inter-task interference by proactively avoiding the allocation of shared, conflicting neurons during the fine-tuning process.

\subsection{Ablation Study}\label{secexablation}

To elucidate the intrinsic contributions of our proposed framework, we compare our method against four distinct localization strategies: 

(1) \textbf{Full-Param}: Standard full-parameter fine-tuning without any localization, designed to isolate the overall impact of applying a localization mechanism. 

(2) \textbf{Random}: Random parameter localization that rigorously preserves the exact rank distribution of our method, isolating the effect of merely reducing the trainable parameter count. 

(3) \textbf{Static}: Localization derived directly from the current model parameters (Localization from $\theta$), serving as a baseline for pre-update parameter importance. 

(4) \textbf{Probing}: Localization based on the model post-probing SFT (Localization from $\theta'$), designed to contrast the efficacy of the intermediate state $\theta'$ against our estimated ideal state $\hat{\theta}^I$. 

As presented in Table~\ref{tab:ablation_levels}, the results not only confirm that localization provides critical interpretability-driven guidance, but also conclusively demonstrate that anticipating "future" parameter states offers significantly greater foresight than relying on current parameters. Additionally, Appendix~\ref{app:robustness_ablation} investigates the robustness of our framework against diverse stochasticity.

\begin{wraptable}{r}{0.52\textwidth}
    \vspace{-10pt} 
    \centering
    \scriptsize
    \renewcommand{\arraystretch}{1.1}
    \setlength{\tabcolsep}{3pt} 
    \resizebox{\linewidth}{!}{
    \begin{tabular}{lcccccc}
        \toprule
        \multirow{2}{*}{\textbf{Strategy}} & \multicolumn{3}{c}{\textbf{Neuron-Level}} & \multicolumn{3}{c}{\textbf{Component-Level}} \\
        \cmidrule(lr){2-4} \cmidrule(lr){5-7}
        & \textbf{LR} & \textbf{NLU} & \textbf{MR} & \textbf{LR} & \textbf{NLU} & \textbf{MR} \\
        \midrule
        Full-Param & 69.27 & 94.03 & 15.66 & 99.73 & 96.33 & 71.83 \\
        Random     & 88.31 & 95.03 & 43.77 & 99.64 & 96.86 & 76.41 \\
        Static     & 87.49 & 95.52 & 55.83 & 99.75 & 96.33 & 83.52 \\
        Probing    & 92.76 & 96.10 & 87.16 & 99.96 & 96.21 & 99.60 \\
        \midrule
        \textbf{Ours} & \textbf{98.51} & \textbf{96.31} & \textbf{98.79} & \textbf{100.00} & \textbf{97.55} & \textbf{100.00} \\
        \bottomrule
    \end{tabular}
    }
    \vspace{-5pt} 
    \caption{Ablation study across datasets at neuron and component levels.}
    \label{tab:ablation_levels}
    \vspace{-10pt} 
\end{wraptable}

\begin{figure}[htbp]
    \centering
    
    \begin{subfigure}[b]{0.48\textwidth}
        \centering
        \includegraphics[width=\textwidth]{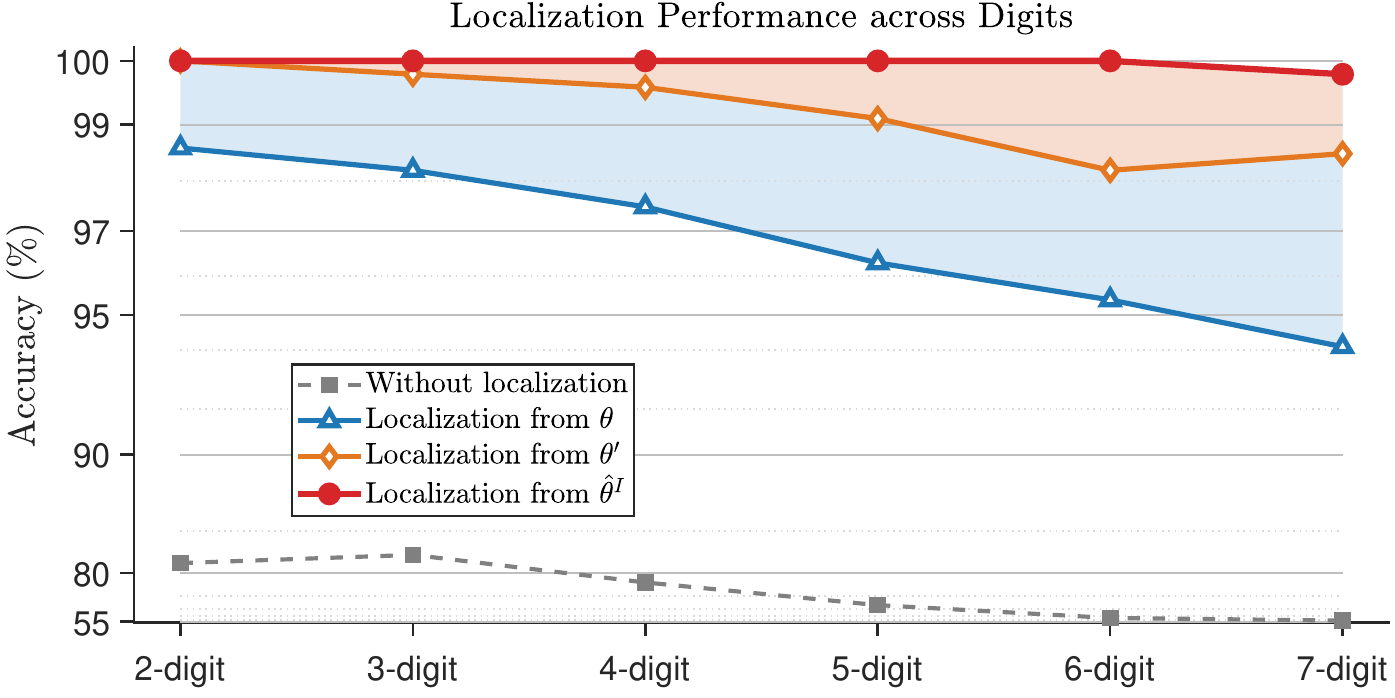}
        \caption{Ablation on different digits.}
        \label{fig:ablation_digit}
    \end{subfigure}
    \hfill 
    \begin{subfigure}[b]{0.48\textwidth}
        \centering
        \includegraphics[width=\textwidth]{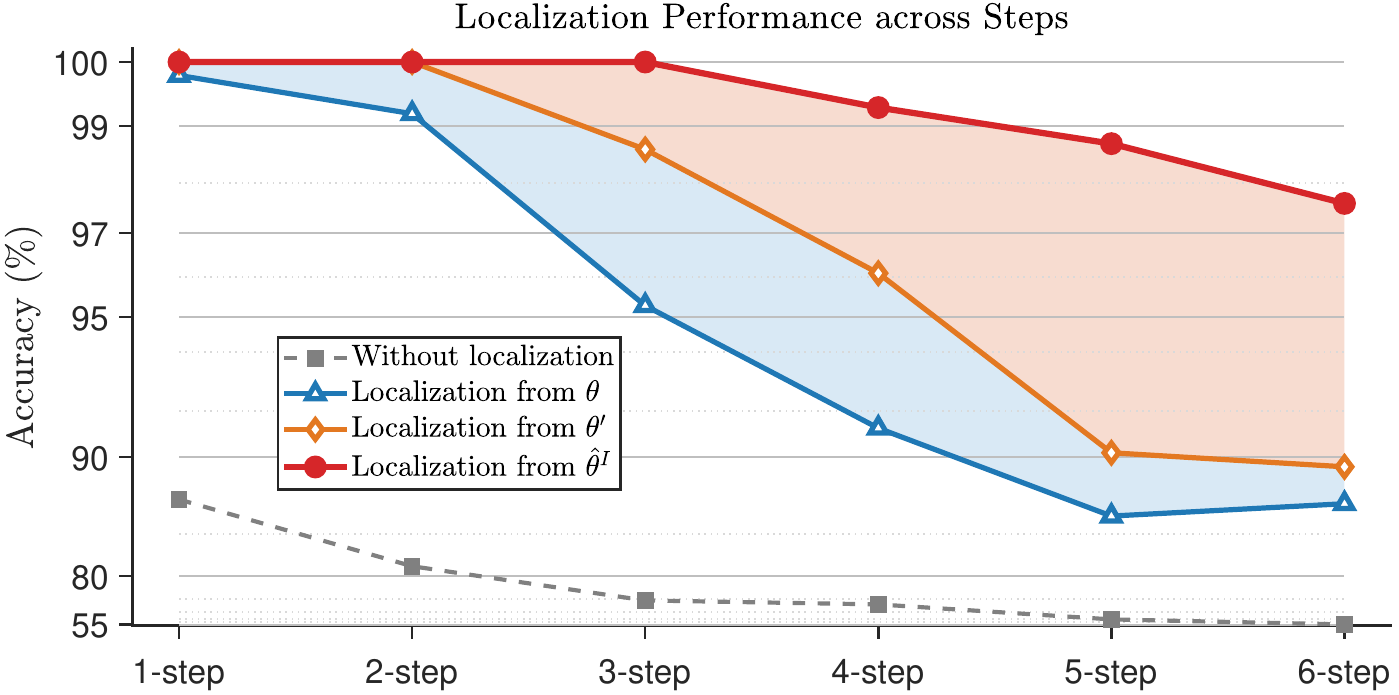}
        \caption{Ablation on different steps.}
        \label{fig:ablation_step}
    \end{subfigure}
    
    \caption{Ablation study results on the arithmetic task. (a) Performance variations across different digits. (b) Performance variations across different reasoning steps.}
    \label{fig:arithmetic_ablation_main}
\end{figure}

Although the Probing strategy performs comparably to our method on standard benchmarks, we conduct further evaluations on tasks of escalating difficulty to explicitly disentangle their capabilities. Specifically, we utilize Arithmetic subtasks ranging from 2 to 7 digits and 1 to 6 reasoning steps. As illustrated in Figure~\ref{fig:arithmetic_ablation_main}, the performance of both the Static and Probing strategies degrades precipitously as task complexity increases, whereas our method preserves substantial robustness. This phenomenon occurs because harder tasks necessitate significantly larger parameter updates during fine-tuning, thereby exacerbating the representational lag inherent in both the initial parameters $\theta$ and the probing state $\theta'$. These findings substantiate that our method is uniquely advantageous for complex tasks where the model struggles to rapidly adapt. Conversely, for rudimentary SFT tasks under extreme computational constraints, the Probing or Static methods may serve as viable, efficient alternatives.

Furthermore, Appendix~\ref{suppMI} provides an in-depth mechanistic interpretability analysis. From a circuit perspective, this analysis corroborates that our estimated parameters yield a mechanistic circuit more closely aligned with that of the fully fine-tuned model, significantly outperforming direct localization from either the initial state $\theta$ or the probing state $\theta'$. Additionally, Appendix~\ref{suppgraph} visualizes the respective circuit graphs for all three parameter states: $\theta$, $\theta'$, and $\hat{\theta}^{I}$. These visualizations explicitly demonstrate that our $K$-value extrapolation successfully identifies critical "intermediate nodes." These intermediate nodes typically encode distinct subskills within the computational pathways, constituting essential components of the underlying task mechanism.

\subsection{Analysis about Probing SFT}\label{secexprobing}
\begin{figure*}[htbp]
    \centering
    \includegraphics[width=0.96\textwidth]{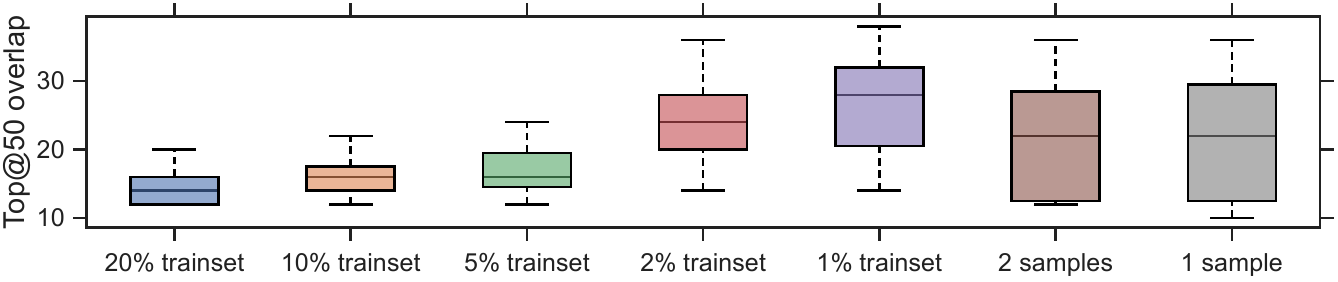}
    \caption{Top 50 overlap with Full-Parameter fine-tuning model across different probing SFT setups. }
    \label{fig:differentprobingsetup}
\end{figure*}

In this section, we investigate the empirical impact of various probing SFT setups. Using the fully fine-tuned model as a surrogate for the ideal model, we adopt the component-level estimation pipeline on the Mistral-7B model for the IOI task as our evaluation baseline. We systematically compare probing models updated on varying proportions of the training dataset: 20\%, 10\%, 5\%, 2\%, and 1\%, alongside extreme scenarios restricted to merely 2 samples and 1 sample. For each configuration, we run 10 random seeds and sample 50 distinct $K$ values per run within a valid range and compute the Top@50 overlap between the component rankings of our estimated $\hat{\theta}^{I}$ and those of the surrogate ideal model.

As illustrated in Figure~\ref{fig:differentprobingsetup}, reducing the sample size used for the probing SFT consistently improves the mean overlap with the ideal model up to a certain threshold. Moreover, we also validate that the learning rate and epoch expose the similar pattern in Appendix~\ref{supplearningfactor}. These empirical trends corroborate our theoretical assertion in Section~\ref{secprobingsft}: maintaining a sufficiently small parameter update during probing is imperative to preserve the necessary degrees of freedom for the extrapolation term $K \cdot \Delta \theta$. However, in extreme low-data regimes (i.e., 1 or 2 samples), the variance of the estimations escalates sharply, precipitating a noticeable decline in the mean overlap. We attribute this degradation to the inability of such minuscule sample sets to adequately capture the underlying distribution of the target task, which inadvertently injects significantly biased gradient directions into $\Delta \theta$. 

\section{Conclusion and Limitations}
\label{sec:conclusion}

This paper presents a breakthrough advancement in mechanistic localization prior to Supervised Fine-Tuning (SFT), enabling the prediction of post-SFT mechanisms to proactively facilitate the tuning process. Theoretically, we bridge mechanistic interpretability with parameter updates via Taylor expansion, breaking the fundamental post-hoc nature of interpretability analysis to usher in a predictive, future-oriented paradigm—thereby transforming interpretability from a retrospective diagnostic tool into an actionable, proactive optimizer. Practically, utilizing only the base model parameters and a probing SFT model trained on a mere $1\%$ of the dataset, we effectively estimate and rank the mechanistic importance of parameters for the anticipated post-SFT state. This enables a targeted, fine-grained allocation of tuning intensity while ensuring scalability in both time and performance. Our work inspires future engineering-oriented interpretability research based on following \textbf{limitations}:

\begin{itemize}
    \item \textbf{Mechanistic Interpretability for Multi-Token Generation:} Current interpretability methods predominantly focus on the internal mechanisms during next-token prediction. However, mainstream post-training SFT typically targets instruction-following tasks that necessitate long-sequence generation~\citep{lai2025survey}. This discrepancy currently restricts the locating-then-tuning paradigm to relatively narrow downstream applications—such as LLM unlearning and knowledge editing—rather than broader instruction task practices. Bridging this gap poses a revolutionary challenge: extracting representative mechanisms and their corresponding parameter regions across multiple forward passes. Exploring this frontier will directly extend our insights into critical domains like reinforcement learning and on-policy distillation.
    
    \item \textbf{Targeted Parameter Updates for Conflicting Mechanisms:} Due to neuron polysemanticity, fine-tuning on a specific task frequently degrades capabilities on others. Existing interpretability and localization efforts merely identify these polysemantic neurons, falling short of providing effective mitigation strategies. This remains a limitation of our study, which renders the advantages of multi-objective joint optimization less pronounced. Addressing this issue will tightly couple mechanistic interpretability with post-training optimization, drastically expanding the practical utility of the interpretability field.
\end{itemize}

\bibliography{iclr2025_conference}
\bibliographystyle{abbrvnat}

\newpage
\appendix
\section{The Derivation of Equation~\ref{eqtsingle}}\label{suppderivationeqt2}
\label{appendix:derivation}

In this appendix, we provide the detailed derivation for Equation~\ref{eqtsingle}, which approximates the causal effect of an infinitesimal parameter update $\theta'$ using terms computed entirely under the current parameters $\theta$.

First, we treat any activation $a$ as a differentiable function of the model parameters $\theta$. Applying a first-order Taylor expansion with respect to $\theta$, we can approximate the clean and corrupted activations under $\theta'$ as follows:
\begin{align}
    A^{\theta'}(x_r) &\approx A^{\theta}(x_r) + \Delta \theta \cdot \frac{\partial A^{\theta}(x_r)}{\partial \theta} \\
    A^{\theta'}(x_c) &\approx A^{\theta}(x_c) + \Delta \theta \cdot \frac{\partial A^{\theta}(x_c)}{\partial \theta}
\end{align}
Subtracting these two equations provides a linear approximation of the activation difference, separating the original difference from the effect of the parameter update:
\begin{equation}
    A^{\theta'}(x_c) - A^{\theta'}(x_r) \approx (A^{\theta}(x_c) - A^{\theta}(x_r)) + \Delta \theta \cdot \left( \frac{\partial A^{\theta}(x_c)}{\partial \theta} - \frac{\partial A^{\theta}(x_r)}{\partial \theta} \right)
\end{equation}

Next, we approximate the gradient term. Let the gradient under the current parameters $\theta$ be denoted as $g(\theta) = \frac{\partial f^{\theta}_a}{\partial a^{\theta}}$. Applying a first-order Taylor expansion to the gradient for the updated parameters $\theta'$ yields:
\begin{equation}
    g(\theta') \approx g(\theta) + \Delta \theta \cdot \frac{\partial g(\theta)}{\partial \theta}
\end{equation}
By substituting the definition of $g(\theta)$ into the derivative term, we can further expand it into a mixed second-order derivative:
\begin{equation}
    \frac{\partial g(\theta)}{\partial \theta} = \frac{\partial}{\partial \theta} \left( \frac{\partial f^{\theta}_a}{\partial a^{\theta}} \right) = \frac{\partial^2 f^{\theta}_a}{\partial a^{\theta} \partial \theta}
\end{equation}

Finally, we substitute both the activation difference approximation and the gradient approximation back into the Attribution Patching formulation for $\theta'$. By expanding the product and omitting higher-order terms $\mathcal{O}((\Delta \theta)^2)$, we arrive at our final formulation:
\begin{align*}
    f^{\theta'}_a(A(x_c)) - f^{\theta'}_a(A(x_r)) &\approx \left(A^{\theta'}(x_c) - A^{\theta'}(x_r)\right) \cdot \frac{\partial f^{\theta'}_a}{\partial a^{\theta'}}\bigg|_{a=A^{\theta'}(x_r)} \\
    &\approx \left[ \left(A^{\theta}(x_c) - A^{\theta}(x_r)\right) + \Delta \theta \cdot \left( \frac{\partial A^{\theta}(x_c)}{\partial \theta} - \frac{\partial A^{\theta}(x_r)}{\partial \theta} \right) \right] \\
    &\quad \cdot \left[ g(\theta) + \Delta \theta \cdot \frac{\partial^2 f^{\theta}_a}{\partial a^{\theta} \partial \theta} \right] \\
    &= \left(A^{\theta}(x_c) - A^{\theta}(x_r)\right) \cdot \frac{\partial f^{\theta}_a}{\partial a^{\theta}} + \Delta \theta \cdot \left(A^{\theta}(x_c) - A^{\theta}(x_r)\right) \cdot \frac{\partial^2 f^{\theta}_a}{\partial a \partial \theta} \\
    &\quad + \Delta \theta \cdot \left( \frac{\partial A^{\theta}(x_c)}{\partial \theta} - \frac{\partial A^{\theta}(x_r)}{\partial \theta} \right) \cdot \frac{\partial f^{\theta}_a}{\partial a^{\theta}} + \mathcal{O}((\Delta \theta)^2) \\
    &\approx f^{\theta}_a(A(x_c)) - f^{\theta}_a(A(x_r)) \\
    &\quad + \Delta \theta \cdot \left[ \left( \frac{\partial A^{\theta}(x_c)}{\partial \theta} - \frac{\partial A^{\theta}(x_r)}{\partial \theta} \right) \cdot \frac{\partial f^{\theta}_a}{\partial a^{\theta}} + \left(A^{\theta}(x_c) - A^{\theta}(x_r)\right) \cdot \frac{\partial^2 f^{\theta}_a}{\partial a \partial \theta} \right] \\
    &= f^{\theta}_a(A(x_c)) - f^{\theta}_a(A(x_r)) + \Delta \theta \cdot S(\theta)
\end{align*}
where $S(\theta)$ effectively captures the sensitivity of the target neuron $a$'s attribution value to the parameter updating.

\section{Derivation of the Translation Operator for Sensitivity $S(\theta)$}
\label{appendix:translation}

In this section, we provide the mathematical proof demonstrating how the sensitivity at the ideal state, $S(\hat{\theta}^I)$, can be evaluated as $S(\theta + K \Delta \theta)$ by circumventing the intractable higher-order derivatives.

\paragraph{Simplifying the Notation and Revealing the Identity.}
Let $\Delta A^\theta = A^\theta(x_c) - A^\theta(x_r)$. According to Equation~3, the sensitivity $S(\theta)$ can be rewritten as:
\begin{equation*}
    S(\theta) = \frac{\partial \Delta A^\theta}{\partial \theta} \cdot \frac{\partial f^\theta_a}{\partial a^\theta} + \Delta A^\theta \cdot \frac{\partial^2 f^\theta_a}{\partial a \partial \theta}
\end{equation*}
Applying the product rule in reverse reveals that $S(\theta)$ is strictly equivalent to the gradient of the attribution score with respect to the parameters:
\begin{equation*}
    S(\theta) \equiv \nabla_{\theta} \left[ \Delta A^{\theta} \cdot \frac{\partial f^{\theta}_a}{\partial a^{\theta}} \right] \equiv \nabla_{\theta} [\Delta E^{\theta}_{\mathcal{D}_{\mathcal{T}}}(a)]
\end{equation*}
This insight indicates that our iterative update formula, $\Delta E^{\theta^{k+1}}_{\mathcal{D}_{\mathcal{T}}}(a) = \Delta E^{\theta^{k}}_{\mathcal{D}_{\mathcal{T}}}(a) + \Delta \theta^k \cdot S(\theta^k)$, is fundamentally solving an Ordinary Differential Equation (ODE) via Euler's Method.

\paragraph{The Differential Operator and Infinite Series.}
Assuming we perform infinite micro-updates along a fixed direction $\Delta \theta$, we define a differential operator $D$:
\begin{equation*}
    D = \Delta \theta \cdot \nabla_{\theta}
\end{equation*}
which computes the directional derivative along $\Delta \theta$. Utilizing the Taylor expansion, the sensitivity at step $k$ unfolds into an infinite series:
\begin{equation*}
    S(\theta^k) = S(\theta^0) + k D S(\theta^0) + \frac{k^2}{2!} D^2 S(\theta^0) + \frac{k^3}{3!} D^3 S(\theta^0) + \dots
\end{equation*}
Notice that the operators $D^2, D^3, \dots$ correspond exactly to the computationally intractable second-order Hessians, third-order tensors, and beyond.

\paragraph{Operator Folding and Translation.}
In functional analysis, this infinite series perfectly matches the Taylor expansion of $e^x$. We can rigorously fold it into an exponential operator:
\begin{equation*}
    S(\theta^k) = \left( \sum_{n=0}^{\infty} \frac{(k D)^n}{n!} \right) S(\theta^0) = e^{k \Delta \theta \cdot \nabla_{\theta}} S(\theta^0)
\end{equation*}
In Lie Algebra~\citep{jacobson2013lie} and Quantum Mechanics~\citep{messiah2014quantum}, $e^{v \cdot \nabla}$ represents the Translation Operator, which systematically shifts the independent variable of any analytic function: $e^{v \cdot \nabla} f(x) \equiv f(x + v)$. Applying this property to our exponential operator yields:
\begin{equation*}
    e^{k \Delta \theta \cdot \nabla_{\theta}} S(\theta^0) \equiv S(\theta^0 + k \Delta \theta) \implies S(\theta^k) = S(\theta^0 + k \Delta \theta)
\end{equation*}
Consequently, for the final ideal state $I$ (reached after $K$ steps), we arrive at $S(\theta^I) = S(\theta^0 + K \Delta \theta)$. This elegant identity demonstrates that to evaluate the sensitivity at the ideal state, we merely need to scale the step distance $K$, completely avoiding the computation of higher-order derivative terms.

\section{Why the Probing Model's Attribution Cannot Directly Substitute the Ideal Model}
\label{suppproofprobing}

In this appendix, we demonstrate why it is theoretically flawed to directly substitute the attribution patching value of the probing model, denoted here as an arbitrary discrete step $\theta^k$, for that of the ideal model $\theta^I$. That is, we prove why the approximation $\Delta E^{\theta^{I}}_{\mathcal{D}_{\mathcal{T}}}(a) \approx \Delta E^{\theta^k}_{\mathcal{D}_{\mathcal{T}}}(a)$ does not generally hold.

To simplify the notation, let $X_0 = A^{\theta^0}(x_c) - A^{\theta^0}(x_r)$ and $Y_0 = \frac{\partial f^{\theta^0}_a}{\partial a^{\theta^0}}$ represent the activation difference and gradient for the current model $\theta^0$, respectively. Similarly, let $X_k = A^{\theta^k}(x_c) - A^{\theta^k}(x_r)$ and $Y_k = \frac{\partial f^{\theta^k}_a}{\partial a^{\theta^k}}$ represent the corresponding terms for the probing model at step $k$.

Recall our causal effect estimation formulation:
\begin{equation}
    \Delta E^{\theta^{I}}_{\mathcal{D}_{\mathcal{T}}}(a) \approx X_0 Y_0 + \frac{1}{2} \left[ S(\theta^0) + S(\theta^k) \right] \cdot \Delta \theta
\end{equation}

Expanding the sensitivity term $S(\theta^0) \cdot \Delta \theta$ yields:
\begin{equation}
    S(\theta^0) \cdot \Delta \theta = \left( \frac{\partial X_0}{\partial \theta} \cdot \Delta \theta \right) \cdot Y_0 + X_0 \cdot \left( \frac{\partial Y_0}{\partial \theta} \cdot \Delta \theta \right)
\end{equation}
At this juncture, we introduce the \emph{HVP Assumption} (which will be detailed in Section~4.2), allowing us to approximate the gradient variation as $\frac{\partial Y_0}{\partial \theta} \cdot \Delta \theta \approx Y_k - Y_0$. Applying this assumption, we obtain:
\begin{equation}
    \label{eq:c1}
    S(\theta^0) \cdot \Delta \theta \approx \left( \frac{\partial X_0}{\partial \theta} \cdot \Delta \theta \right) \cdot Y_0 + X_0(Y_k - Y_0)
\end{equation}
By symmetric logic, for the probing state $\theta^k$:
\begin{equation}
    \label{eq:c2}
    S(\theta^k) \cdot \Delta \theta \approx \left( \frac{\partial X_k}{\partial \theta^k} \cdot \Delta \theta \right) \cdot Y_k + X_k(Y_k - Y_0)
\end{equation}

The equivalence $\Delta E^{\theta^{I}}_{\mathcal{D}_{\mathcal{T}}}(a) \approx \Delta E^{\theta^k}_{\mathcal{D}_{\mathcal{T}}}(a)$ can only be established if we impose an extremely strong assumption regarding the activation dynamics:
\begin{equation}
    \frac{\partial (A(x_c) - A(x_r))}{\partial \theta} \cdot \Delta \theta \approx X_k - X_0
\end{equation}
\emph{If and only if} this assumption holds, we can substitute it into Equation~\ref{eq:c1} and Equation~\ref{eq:c2}:
\begin{align}
    S(\theta^0) \cdot \Delta \theta &\approx (X_k - X_0)Y_0 + X_0(Y_k - Y_0) = X_k Y_0 + X_0 Y_k - 2X_0 Y_0 \\
    S(\theta^k) \cdot \Delta \theta &\approx (X_k - X_0)Y_k + X_k(Y_k - Y_0) = 2X_k Y_k - X_0 Y_k - X_k Y_0
\end{align}
Summing these two refined equations together provides:
\begin{equation}
    \frac{1}{2} \left[ S(\theta^0) + S(\theta^k) \right] \cdot \Delta \theta \approx \frac{1}{2} [2X_k Y_k - 2X_0 Y_0] = X_k Y_k - X_0 Y_0
\end{equation}
Finally, substituting this result back into our main formulation yields:
\begin{equation}
    \Delta E^{\theta^{I}}_{\mathcal{D}_{\mathcal{T}}}(a) \approx X_0 Y_0 + X_k Y_k - X_0 Y_0 = X_k Y_k = \Delta E^{\theta^k}_{\mathcal{D}_{\mathcal{T}}}(a)
\end{equation}

This derivation reveals that directly using the probing model's attribution relies on the assumption that the activation differences change perfectly linearly with the parameter update across the macroscopic step from $\theta^0$ to $\theta^k$. Because deep neural networks are highly non-linear, this stringent constraint is almost never satisfied during discrete gradient updates, rendering the naive substitution $\Delta E^{\theta^{I}}_{\mathcal{D}_{\mathcal{T}}}(a) \approx \Delta E^{\theta^k}_{\mathcal{D}_{\mathcal{T}}}(a)$ theoretically unsound.

\section{Details of Datasets}\label{suppdetails}
To facilitate the implementation of mechanistic interpretability, we strictly constrain the label of each dataset to a single token. Consequently, for specific datasets, we reformulate the inputs into question-answering prompt templates. The Arithmetic dataset is systematically partitioned into sub-datasets based on \emph{digit} and \emph{step} complexity, encompassing solely the four fundamental operations (addition, subtraction, multiplication, and division). Here, \emph{digit} denotes the number of digits in the participating operands, while \emph{step} indicates the number of mathematical operators within a single sample (excluding the equals sign). The statistical distribution and prompt examples for all datasets are detailed in Table~\ref{tab:dataset_details}.

\begin{table}[htbp]
    \centering
    \scriptsize
    \renewcommand{\arraystretch}{1.15}
    \resizebox{1\textwidth}{!}{
    \begin{tabular}{@{}ll|ccc p{1.3cm} p{6.5cm} p{1.2cm}@{}}
        \toprule
        \multicolumn{2}{@{}l|}{\textbf{Dataset}} & \textbf{Train} & \textbf{Test} & \textbf{Max Len} & \textbf{Output} & \textbf{Input Prompt Case} & \textbf{Label} \\
        \midrule
        \multicolumn{2}{@{}l|}{IOI} & 6,000 & 600 & 50 & Correct \newline name & 
        Then, Aaron and Amber were working at the house. Aaron decided to give a drink to & Amber \\
        \midrule
        \multicolumn{2}{@{}l|}{Gender} & 3,000 & 3,000 & 50 & he / she & 
        So Kelly is always up for an adventure, isn't & she \\
        \midrule
        \multicolumn{2}{@{}l|}{BOOL} & 10,000 & 4,000 & 100 & True / \newline False & 
        Evaluate the following boolean expression as either 'True' or 'False'. \newline ( not ( True or True ) ) and ( False or ( not True ) ) & False \\
        \midrule
        \multicolumn{2}{@{}l|}{SST-2} & 67,349 & 872 & 200 & positive / \newline negative & 
        Is the sentiment of following sentence positive or negative? that 's far too tragic to merit such superficial treatment. Answer: It is & negative \\
        \midrule
        \multicolumn{2}{@{}l|}{MRPC} & 3,670 & 1,730 & 200 & Yes / \newline No & 
        Are the meanings of the following two sentences equivalent? 1. And if both apply, they are essentially possible. 2. And if both apply, they are essentially impossible. & No \\
        \midrule
        \multicolumn{2}{@{}l|}{QQP} & 364,000 & 391,000 & 200 & Yes / \newline No & 
        Is Question ``How do you control your horniness?'' a duplicate of Question ``How do I control my horny emotions?''? & Yes \\
        \midrule
        \multicolumn{2}{@{}l|}{MNLI} & 393,000 & 9,800 & 250 & Option \newline Letter & 
        What is the relationship of premise ``How do you know? All this is their information again.'' to hypothesis ``This information belongs to them.'': A. entailment, B. contradiction, or C. unrelated? The answer is & A \\
        \midrule
        \multicolumn{2}{@{}l|}{RTE} & 2,490 & 3,000 & 200 & Yes / \newline No & 
        Does sentence ``Oil prices fall back as Yukos oil threat lifted'' entail sentence ``Oil prices rise.''? & Yes \\
        \midrule
        \multicolumn{2}{@{}l|}{WinoGrande} & 9,248 & 1,267 & 200 & Yes / No & 
        The trophy doesn't fit into the brown suitcase because \_ is too large. Should the '\_' be the trophy? & Yes \\
        \midrule
        \multicolumn{2}{@{}l|}{Docstring} & 2,400 & 600 & 50 & Missing \newline param & 
        def process(self, data, name, value, result, line, file): \newline """control action cost \newline :param value: research subject \newline :param result: health issue \newline :param & line \\
        \midrule
        \multicolumn{2}{@{}l|}{Induction} & 2,400 & 600 & 600 & Last token & 
        After Gray applied for health insurance, Alice applied & for \\
        \midrule
        \multirow{6}{*}{Arithmetic} 
        & 2-digit & 10,000 & 600 & 160 & Option \newline Letter & 
        Please choose the correct option from the following: "What is 80 plus 29?" Options: A. \#\#\#\#\#, B. dnfidlg, C. 378, D. 54942, E. 388, F. 109, G. 4. The answer is & F \\
        \cmidrule{2-8}
        & 3-digit & 10,000 & 600 & 160 & Option \newline Letter & 
        Please choose the correct option from the following: "What is 161 plus 390?" Options: A. 16957, B. 551, C. people, D. 908, E. 305, F. 5, G. gkledns. The answer is & B \\
        \cmidrule{2-8}
        & 7-digit & 10,000 & 600 & 160 & Option \newline Letter & 
        Please choose the correct option from the following: "What is 5567345 plus 4592838?" Options: A. 10160183, B. 552493, C. apple, D. 374329826, E. 5567838, F. dhjklng, G. 387439. The answer is & A \\
        \cmidrule{2-8}
        & 1-step & 10,000 & 600 & 160 & Option \newline Letter & 
        Please choose the correct option from the following: "34-62=" Options: A. -28, B. 34, C. from, D. 345, E. 11, F. 63, G. right The answer is & A \\
        \cmidrule{2-8}
        & 2-step & 10,000 & 600 & 160 & Option \newline Letter & 
        Please choose the correct option from the following: "6361-32*537=" Options: A. ill, B. 10823, C. -10823, D. 583567, E. 3562, F. -55, G. same. The answer is & C \\
        \cmidrule{2-8}
        & 6-step & 10,000 & 600 & 160 & Option \newline Letter & 
        Please choose the correct option from the following: "5392/57892+92743-372984/239*4428+3792=" Options: A. -6813812, B. 205637, C. lofdt, D. 3034672, E. 2647, F. 109872, G. muly The answer is & A \\
        \bottomrule
    \end{tabular}}
    \caption{Dataset statistics, configurations, and prompt construction examples.}
    \label{tab:dataset_details}
\end{table}

\section{Experiment of Multi-Task Fine Tuning}\label{suppmultitask}
\begin{figure*}[htbp]
    \centering
    \includegraphics[width=0.96\textwidth]{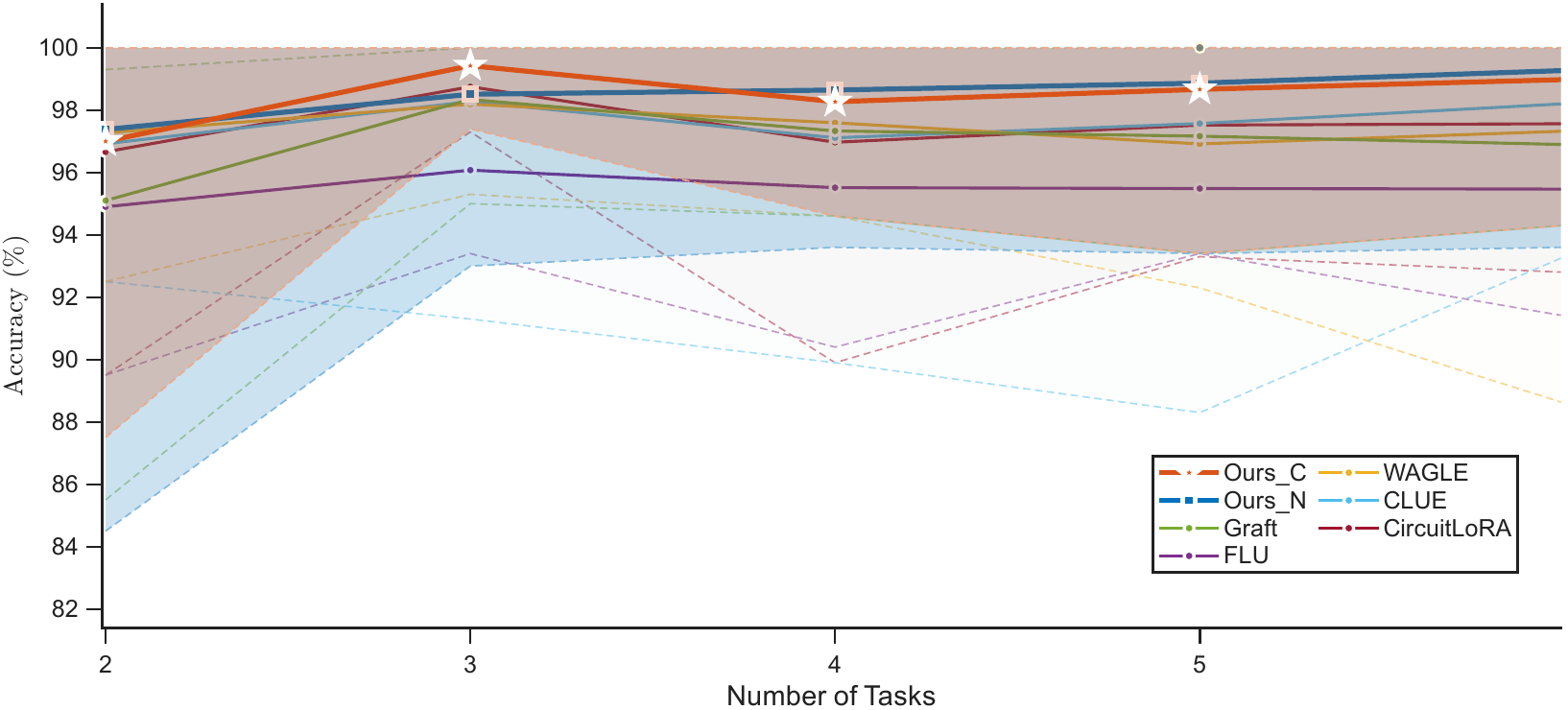}
    \caption{Performance in multi-task accuracy. The solid line represents the mean value, and the dashed line represents the lower bound of the value variation. The shaded areas represent the variances of Our\_C and Our\_N, respectively.}
    \label{fig:accuracy_multitask}
\end{figure*}

\begin{figure*}[htbp]
    \centering
    \includegraphics[width=0.96\textwidth]{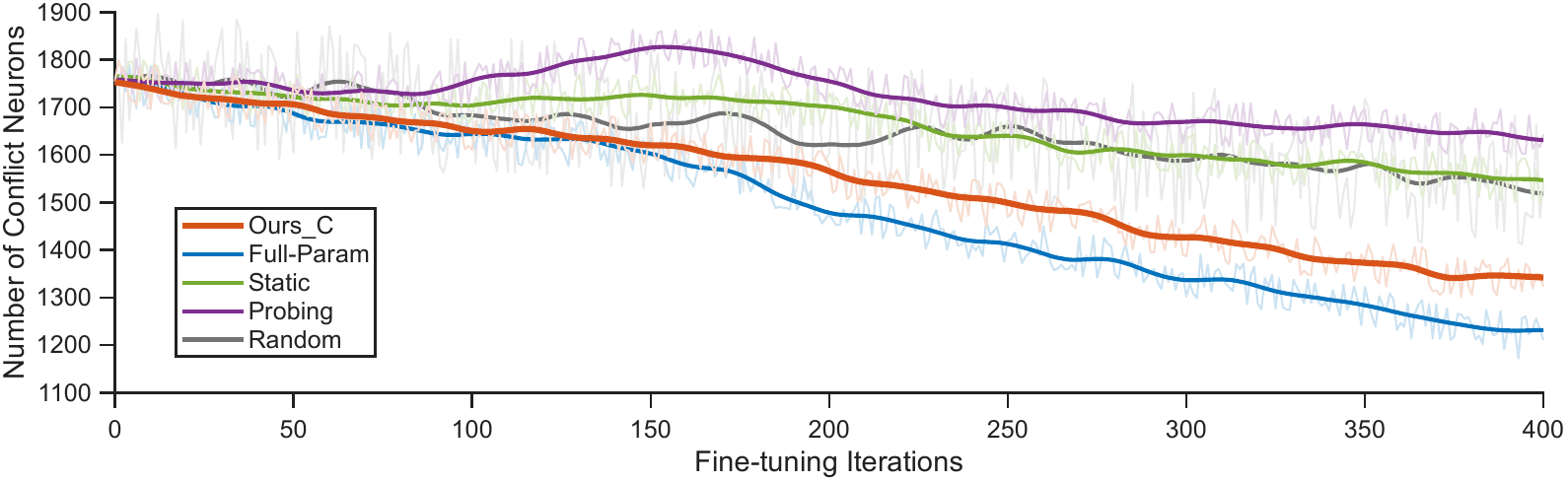}
    \caption{Evolution of conflict component in fine-tuning.}
    \label{fig:conflict_multitask}
\end{figure*}

To verify whether our localization method can mitigate mechanistic inter-task conflicts through forward-looking interpretability insights, we design a multi-task joint fine-tuning setup where such conflicts are both controllable and observable. Sourcing candidate tasks from Appendix~\ref{suppdetails}, we randomly construct joint task sets ranging from $2$ to $6$ tasks, ensuring at least three distinct random combinations for each set size. We evaluate the mean and variance of the Target Task Accuracy (TTA) across these combinations during joint fine-tuning. As illustrated in Figure~\ref{fig:accuracy_multitask}, our method not only achieves the highest average performance in multi-task scenarios but also demonstrates superior stability, evidenced by nearly the lowest variance. This indicates the absence of severe inter-task conflicts that would otherwise cause catastrophic degradation on individual tasks.

To further substantiate this conclusion, we trace the evolution of conflicting nodes within the model circuits across checkpoints from $0$ to $400$ fine-tuning iterations. By utilizing the boolean solver introduced in the CLUE method to quantify these conflicts, Figure~\ref{fig:conflict_multitask} reveals that our approach significantly minimizes the emergence of conflicting nodes. Its efficacy is surpassed only by unconstrained full-parameter fine-tuning, which inherently represents a natural mechanistic evolution with maximum degrees of freedom. Conversely, baseline methods perform comparably to random localization, indicating a lack of localization precision. Consequently, they fail to reduce conflicting nodes, demonstrating an inability to effectively leverage mechanistic interpretability for practical fine-tuning guidance.

\section{Exploration and Experiments about $K$}\label{supprobustofK}
\subsection{The Optimal Sampling Interval of $K$}
\label{sec:experiments_k_interval}

\begin{figure*}[htbp]
    \centering
    \includegraphics[width=0.8\textwidth]{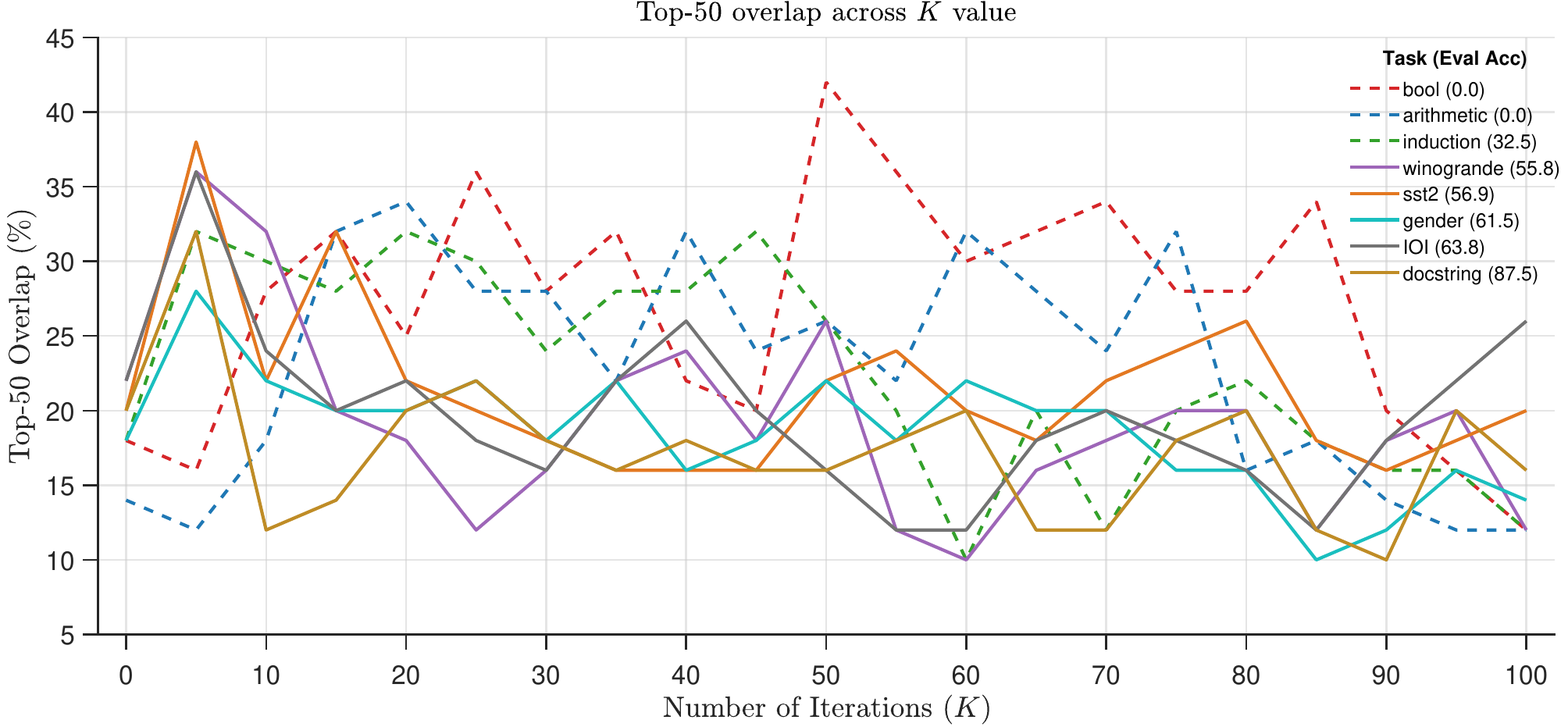}
    \caption{Overlap rate of the top 50 components with the full-parameter circuit ($\mathcal{C}_{\text{Full-Param}}$) across tasks with varying initial accuracies.}
    \label{fig:accuracy_overlap}
\end{figure*}

As discussed in Section~4, our estimation of $\hat{\theta}^{I}$ involves averaging the outcomes of ten randomly sampled values for $K$. Empirically, we observe that each SFT task exhibits a specific optimal sampling interval; sampling $K$ within this range yields a significantly higher expected performance compared to sampling outside of it.

To determine this optimal sampling interval for each task, we utilize the fully fine-tuned model (without any localization or parameter freezing) as a surrogate for the ideal model. This choice is motivated by two factors: (1) The fully fine-tuned model is free from auxiliary parameter constraints, thus most closely approximating the natural trajectory of mechanistic evolution. (2) It guarantees a satisfactory accuracy on the target SFT task, ensuring the formation of a complete and well-defined task mechanism within the model. For brevity, we substitute the ideal model parameters ($\theta^{I}$) with those of the fully fine-tuned model throughout this paper. Subsequently, we uniformly sample $K$ from $0$ to $100$ at intervals of $5$. We then compute the overlap of the top 50 components---ranked by their causal effect---between our estimated $\hat{\theta}^{I}$ and the surrogate $\theta^{I}$. The results are illustrated in Figure~\ref{fig:accuracy_overlap}.

The values in parentheses within the figure legend indicate the initial Target Task Accuracy (TTA) prior to fine-tuning. A clear correlation emerges between this initial TTA and the optimal sampling interval for $K$. For tasks with an initial TTA near zero (e.g., BOOL and Arithmetic), the expected overlap remains consistently high across a broad range of $K \in [10, 80]$. Conversely, for tasks with a higher initial TTA (e.g., Docstring, IOI, and Gender), the optimal $K$ values are highly concentrated within a narrow band of $[0, 10]$. We attribute this phenomenon to the initial TTA reflecting the completeness of the pre-existing task mechanism within the base model. A lower initial TTA implies a deficient mechanism, necessitating more substantial parameter updates during fine-tuning; hence, a larger $K$ is required to adequately extrapolate and explore the distribution of $\hat{\theta}^{I}$. Conversely, a higher initial TTA indicates that the model mechanism is already largely intact, requiring only minor parameter updates, allowing a smaller $K$ to sufficiently approximate $\hat{\theta}^{I}$.

\subsection{Automated and Reliable Determination of $K$ via Linear Probing}
\label{app:probing_k}

\begin{table}[htbp]
    \centering
    \small
    \resizebox{\textwidth}{!}{
    \begin{tabular}{lccccccccc}
        \toprule
        \multirow{2}{*}{\textbf{Method / Strategy}} & \multicolumn{3}{c}{\textbf{Bool}} & \multicolumn{3}{c}{\textbf{SST-2}} & \multicolumn{3}{c}{\textbf{Arithmetic}} \\
        \cmidrule(lr){2-4} \cmidrule(lr){5-7} \cmidrule(lr){8-10}
        & \textbf{$K$} & \textbf{Top@50} & \textbf{TTA} & \textbf{$K$} & \textbf{Top@50} & \textbf{TTA} & \textbf{$K$} & \textbf{Top@50} & \textbf{TTA} \\
        \midrule
        1-Sample Uniform       & 35.6 & 28 & 99.64 & 2.1 & 26 & 96.34 & 71.5 & 30 & 99.51 \\
        10-Sample Average      & [10, 80] & 30.4 & 100.00 & [1, 10] & 29.6 & 97.55 & [10, 80] & 31.4 & 100.00 \\
        \midrule
        Probe (First Layer)    & 15.3 & 24 & 98.82 & 8.73 & 28 & 95.17 & 24.8 & 26 & 99.17 \\
        Probe (Middle Layer)   & 67.2 & 32 & 99.67 & 1.2 & 34 & 96.72 & 72.4 & 30 & 99.88 \\
        Probe (Last Layer)     & 58.5 & 34 & 99.83 & 1.2 & 30 & 97.26 & 77.2 & 32 & 100.00 \\
        \bottomrule
    \end{tabular}
    }
    \caption{Ablation study on determining the extrapolation scalar $K$ via linear probing across different residual stream depths on the Mistral-7B backbone, evaluated against standard sampling baselines.}
    \label{tab:probing_k_estimation}
\end{table}
In our primary locating-then-tuning pipeline, the step scalar $K$ is estimated by averaging across 10 randomly sampled values within a plausible interval. To eliminate the computational overhead of repeated samplings and mitigate heuristic bias in scenarios where a surrogate fully fine-tuned model is unavailable, we investigate an automated probing-based framework to reliably predict $K$ prior to full SFT.
Specifically, we construct a lightweight, two-layer linear probing head and attach it to the residual stream of the Mistral-7B backbone at three representative depths: the \emph{First Layer} (Layer 1), the \emph{Middle Layer} (Layer 16), and the \emph{Last Layer} (Layer 32). The probing head is trained using the Top@50 circuit overlap against the surrogate ideal model as the supervised optimization target.

Table~\ref{tab:probing_k_estimation} compares the proposed probing strategy against standard heuristic baselines (1-Sample and 10-Sample sampling) across the Bool, SST-2, and Arithmetic benchmarks in terms of predicted $K$ values, Top@50 circuit overlap, and downstream Target Task Accuracy (TTA). 

Our empirical observations yield two key insights: 1. The probing model accurately predicts the task-specific $K$ value prior to tuning. It matches or even surpasses the circuit overlap and TTA achieved by the 10-sample averaging method, successfully circumventing the tedious time and computational expenditure induced by multi-round sampling.
2. Deploying the probe at the middle or deep residual stream yields significantly superior predictions compared to the first layer. This discrepancy stems from the representational hierarchy in LLMs: activations in the shallowest residual stream predominantly capture token-level lexical and low-level syntactic information, lacking the high-level semantic abstractions necessary to discern whether the base model already possesses relevant downstream task mechanisms. 

\subsection{Robustness of Sampling $K$}
\begin{figure}[htbp]
    \centering
    
    \begin{subfigure}[b]{0.48\textwidth}
        \centering
        \includegraphics[width=\textwidth]{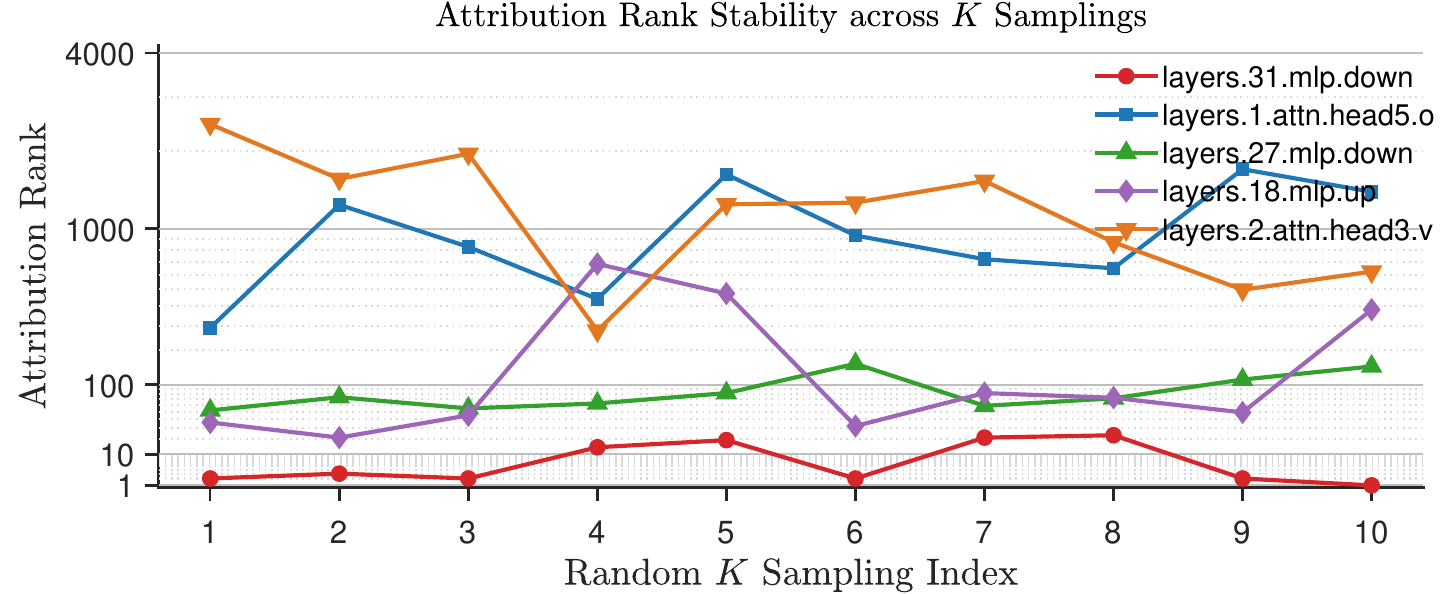}
        \caption{Rank results of 10 samplings for sample nodes.}
        \label{fig:sampling_rank}
    \end{subfigure}
    \hfill
    \begin{subfigure}[b]{0.48\textwidth}
        \centering
        \includegraphics[width=\textwidth]{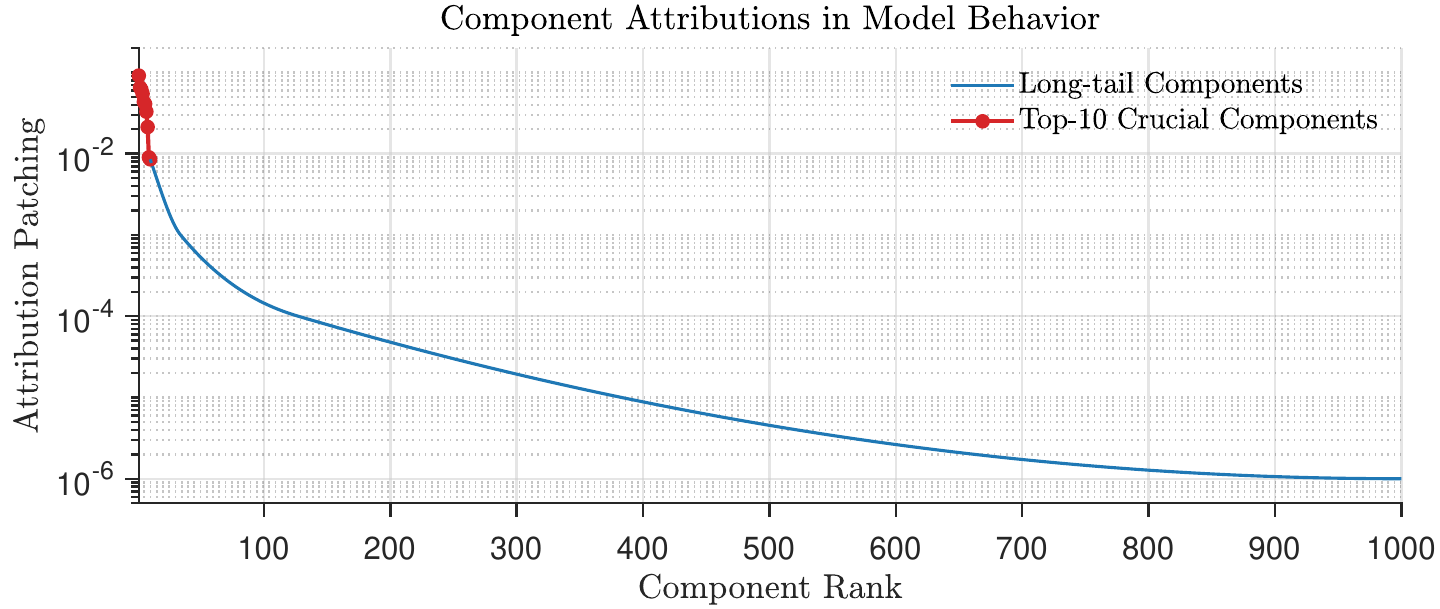}
        \caption{Distribution of attribution value estimates.}
        \label{fig:attribution_dist}
    \end{subfigure}
    
    \caption{Robustness analysis of K-value sampling.}
    \label{fig:k_value_robustness}
\end{figure}

To analyze the stability of our $K$-value sampling, we select a subset of components and track their rank variations across ten independent samplings, as depicted in Figure~\ref{fig:sampling_rank}. The average rankings of these sampled components span a wide spectrum, ranging from the top 10 to beyond 2000. Empirical observations indicate that top-ranked components exhibit minimal rank variance across different samplings. 

Recall that the estimated causal effect $\Delta E^{\theta^{I}}_{\mathcal{D}_{\mathcal{T}}}(a)$ comprises a baseline term $\Delta E^{\theta^0}_{\mathcal{D}_{\mathcal{T}}}(a)$ and a $K$-dependent extrapolation term (i.e., $\frac{1}{2} \left[ S(\theta^0) + S(K \cdot \Delta \theta) \right] \cdot K \cdot \Delta \theta$). By examining the exact magnitudes of these two terms, we uncover an intriguing phenomenon: even for the highest-ranked components, the baseline term $\Delta E^{\theta^0}_{\mathcal{D}_{\mathcal{T}}}(a)$ remains consistently at least one order of magnitude smaller than the $K$-dependent term. This implies that the observed rank stability is not dominated by the intrinsic causal effect of the current model ($\theta$). Instead, the robust gradient direction dictates substantial causal effects across varying $K$ values, precisely isolating the critical nodes that govern the target task. 

Furthermore, Figure~\ref{fig:attribution_dist} illustrates the magnitude distribution of the top 1000 components, revealing an extreme concentration where the top 10 components account for nearly 90\% of the total causal effect. Consequently, any severe rank fluctuations among lower-ranked components during sampling are mathematically negligible and do not compromise the overall efficacy of the localization.

\section{Robustness and Sensitivity Ablation across Stochastic Factors}
\label{app:robustness_ablation}
\begin{figure*}[htbp]
    \centering
    \includegraphics[width=0.96\textwidth]{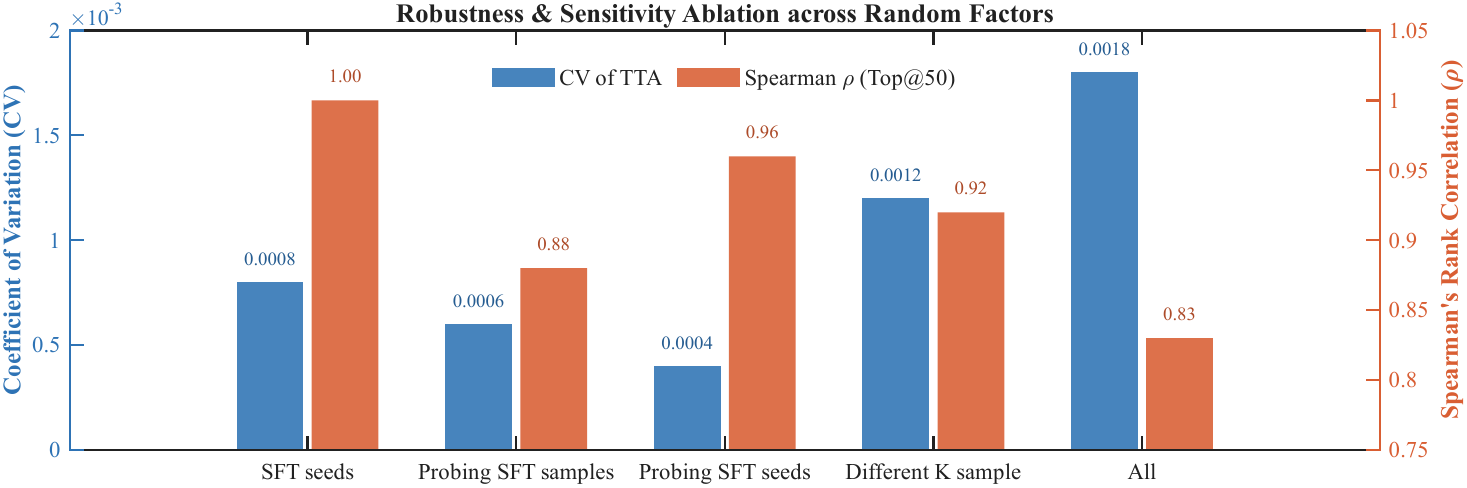}
    \caption{$\rho$ and $CV$ across different stochastic factors. Each set is randomly repeated 10 times on Ours\_C in the Mistral-7B model. }
    \label{fig:robustness_ablation}
\end{figure*}
To rigorously evaluate the algorithmic stability and reproducibility of our framework, we systematically ablate and quantify the impact of key stochastic factors introduced across different operational stages. Specifically, we examine the following five perturbation setups:
\begin{enumerate}
    \item \textbf{SFT Seeds:} Random seeds governing the fine-tuning stage, which control the stochastic initialization of trainable LoRA projection matrices and parameter masking.
    \item \textbf{Probing SFT Samples:} Stochasticity arising from uniformly sub-sampling different $1\%$ data partitions during the probing SFT phase.
    \item \textbf{Probing SFT Seeds:} Random seeds used during probing SFT optimization, which alter the intermediate probing parameter state $\theta'$.
    \item \textbf{Different $K$ Sampling:} Stochasticity induced by independently drawing distinct subsets of the extrapolation step scalar $K$ when estimating $\Delta E^{\theta^{I}}_{\mathcal{D}_{\mathcal{T}}}(a)$.
    \item \textbf{All (Joint Perturbation):} The unconstrained scenario where all four stochastic factors above are simultaneously randomized across runs.
\end{enumerate}

We assess robustness from two complementary perspectives: the intrinsic stability of the identified task circuits and the extrinsic invariance of downstream task performance:
\begin{itemize}
    \item \textbf{Critical Component Stability ($\rho$):} We extract the Top@50 component rankings across independent random runs and compute their pairwise \emph{Spearman's Rank Correlation Coefficient} ($\rho \in [0, 1]$). A correlation value closer to $1$ indicates that the identified computational subgraphs remain highly invariant under random perturbations.
    \item \textbf{Downstream Performance Invariance ($CV$):} We measure the dispersion of the final Target Task Accuracy (TTA) using the \emph{Coefficient of Variation} ($CV$):$CV = \frac{\sigma}{\mu} \times 100\%$, 
    where $\mu$ and $\sigma$ denote the empirical mean and standard deviation of TTA, respectively. A lower $CV$ signifies greater insensitivity of the downstream adaptation to stochastic noise.
\end{itemize}

\paragraph{Empirical Analysis and Insights.}
As illustrated in Figure~\ref{fig:robustness_ablation}, the proposed framework exhibits exceptional resilience across all evaluated dimensions. The downstream performance maintains near-zero variance across all isolated and joint perturbation settings ($CV \le 0.0018$), while the structural ranking of the Top@50 critical components exhibits consistently high correlation ($\rho \ge 0.83$, reaching $1.00$ under varying SFT seeds). These findings confirm that the primary stochastic factors in our pipeline introduce virtually negligible performance instability. From a mechanistic standpoint, this suggests that under our current localization granularity, downstream task adaptation is predominantly governed by a sparse set of dominant, high-attribution neurons and components. Probing with a small parameter displacement ($\Delta\theta$) reliably captures the principal gradient direction of the loss landscape, enabling the extrapolation term to consistently isolate these core task-governing circuits without being derailed by local stochastic sampling fluctuations.

\section{Analysis from Mechanistic Interpretability}\label{suppMI}

\begin{table}[htbp]
    \centering
    \resizebox{0.8\textwidth}{!}{
    \begin{tabular}{lcccccccc}
        \toprule
        \multirow{2}{*}{\textbf{Method}} & \multicolumn{2}{c}{\textbf{Induction}} & \multicolumn{2}{c}{\textbf{Gender}} & \multicolumn{2}{c}{\textbf{IOI}} & \multicolumn{2}{c}{\textbf{Docstring}} \\
        \cmidrule(lr){2-3} \cmidrule(lr){4-5} \cmidrule(lr){6-7} \cmidrule(lr){8-9}
        & \textbf{Top@50} & \textbf{KL} & \textbf{Top@50} & \textbf{KL} & \textbf{Top@50} & \textbf{KL} & \textbf{Top@50} & \textbf{KL} \\
        \midrule
        $\mathcal{C}_{\text{Full-Param}}$ & 100 & 0 & 100 & 0 & 100 & 0 & 100 & 0 \\
        $\mathcal{C}_{\text{Static}}$      & 22 & 1.37 & 18 & 1.57 & 20 & 1.72 & 24 & 1.31 \\
        $\mathcal{C}_{\text{Probing}}$     & 18 & 1.29 & 22 & 2.23 & 22 & 1.25 & 22 & 1.04 \\
        \midrule
        $\mathcal{C}_{\text{Ours}}$        & 28 & 1.03 & 36 & 1.31 & 32 & 0.93 & 28 & 1.01 \\
        \bottomrule
    \end{tabular}
    }
    \caption{Comparison of circuit overlap (Top@50) and KL divergence against the full-parameter finetuned model across interpretability datasets in component level.}
    \label{tab:circuit_comparison1}
\end{table}

\begin{table}[htbp]
    \centering
    \resizebox{0.75\textwidth}{!}{
    \begin{tabular}{lcccccccc}
        \toprule
        \multirow{2}{*}{\textbf{Method}} & \multicolumn{2}{c}{\textbf{Induction}} & \multicolumn{2}{c}{\textbf{Gender}} & \multicolumn{2}{c}{\textbf{IOI}} & \multicolumn{2}{c}{\textbf{Docstring}} \\
        \cmidrule(lr){2-3} \cmidrule(lr){4-5} \cmidrule(lr){6-7} \cmidrule(lr){8-9}
        & \textbf{Top@50} & \textbf{KL} & \textbf{Top@50} & \textbf{KL} & \textbf{Top@50} & \textbf{KL} & \textbf{Top@50} & \textbf{KL} \\
        \midrule
        $\mathcal{C}_{\text{Full-Param}}$ & 100 & 0 & 100 & 0 & 100 & 0 & 100 & 0 \\
        $\mathcal{C}_{\text{Static}}$      & 20 & 1.55 & 24 & 1.61 & 26 & 1.49 & 20 & 1.52 \\
        $\mathcal{C}_{\text{Probing}}$     & 20 & 1.51 & 18 & 1.72 & 20 & 1.62 & 18 & 1.27 \\
        \midrule
        $\mathcal{C}_{\text{Ours}}$        & 32 & 1.11 & 34 & 1.13 & 34 & 1.21 & 30 & 1.15 \\
        \bottomrule
    \end{tabular}
    }
    \caption{Comparison of circuit overlap (Top@50) and KL divergence against the full-parameter finetuned model across interpretability datasets in neuron level.}
    \label{tab:circuit_comparison2}
\end{table}

To provide a rigorous theoretical validation of the differences between our approach and the static/probing localization methods, we employ Edge Attribution Patching (EAP) to construct mechanistic circuits ($\mathcal{C}$) based on the estimated causal effects. Briefly, these circuits are formulated as directed acyclic graphs (DAGs), where nodes denote computational units (neurons or components) and edges represent the activation flows between them. We evaluate the circuits derived from various localization methods using two metrics: the overlap of the top-50 causal effect nodes with the ideal model, and the Kullback-Leibler (KL) divergence between the output logits of the ideal model's circuit and those generated when strictly activating only the nodes and edges within the identified circuit. Consistent with our previous setups, we adopt the fully fine-tuned Mistral-7B model as the ideal surrogate and conduct analyses across four interpretability tasks. Given that the fully fine-tuned model achieves 100\% accuracy on these specific tasks, its derived circuit serves as a robust ground truth.

Tables~\ref{tab:circuit_comparison1} and~\ref{tab:circuit_comparison2} report the results at the component and neuron levels, respectively. Notably, our method yields circuits most closely aligned with the ideal model, demonstrating its superior capability in accurately predicting the post-fine-tuning mechanistic distribution. Furthermore, the negligible performance gap between the probing and static methods implies that the mechanistic distribution of $\theta'$ remains heavily anchored to the initial state $\theta$, thereby lacking the necessary foresight to anticipate mechanistic shifts during the fine-tuning process.

\begin{figure*}[htbp]
    \centering
    \includegraphics[width=0.85\textwidth]{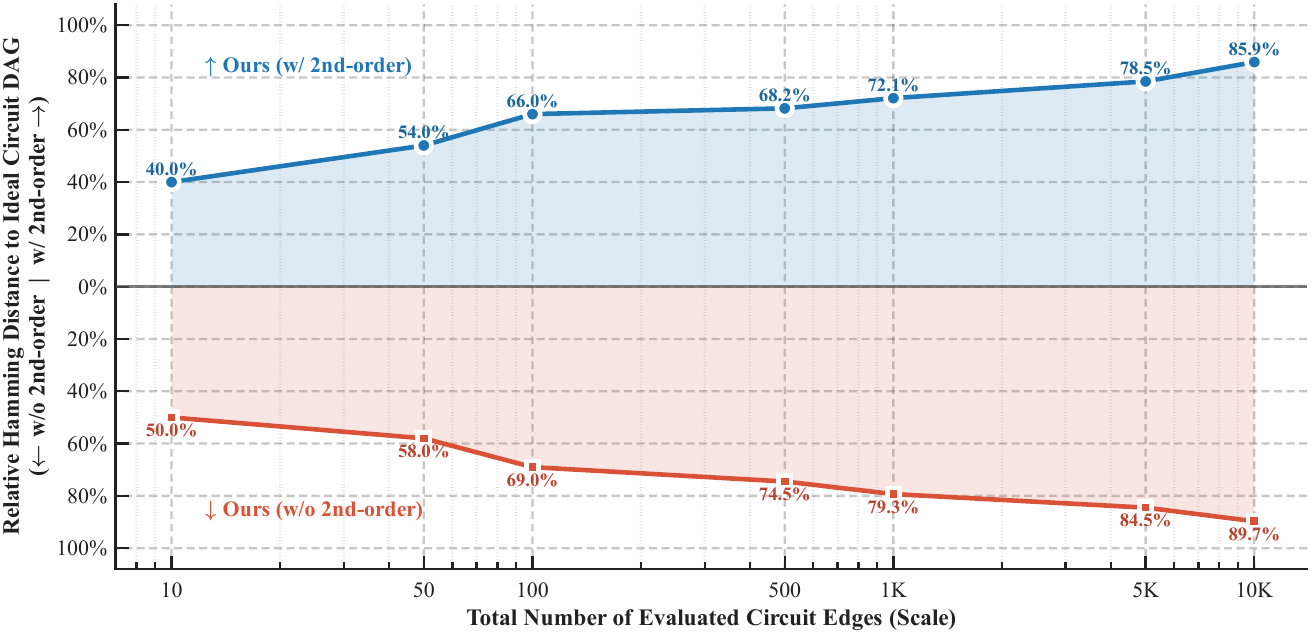}
    \caption{Ablation on Second-Order Approximation across Expanding Edge Capacities}
    \label{fig:circuiterror}
\end{figure*}
Furthermore, to rigorously evaluate the topological fidelity of our predicted circuit graphs and assess the empirical impact of the mixed second-order derivative approximation (Equation~\ref{eqt2order}), we systematically analyze circuit discrepancies against the ideal surrogate model across expanding edge capacities. We measure the \emph{Relative Hamming Distance} (defined as the normalized Hamming distance over the total number of evaluated edges), where a value approaching $100\%$ indicates near-complete divergence between two graph topologies.

As illustrated in Figure~\ref{fig:circuiterror}, while approximating the second-order partial derivative inevitably introduces mild algebraic perturbations, it consistently maintains superior structural fidelity compared to the pure second-order computation across all scales.

Crucially, when synthesized with the attribution magnitude distribution shown in Figure~\ref{fig:attribution_dist}, the empirical efficacy of our framework is overwhelmingly governed by the accurate and robust ranking of the most prominent components (approximately within the Top@50). As established in our attribution analysis, these top-tier computational units account for the vast majority (nearly $90\%$) of the total causal effect governing task adaptation.

\section{Circuit Graph Visualizations}\label{suppgraph}

\begin{figure*}[htbp]
    \centering
    
    \begin{subfigure}{\textwidth}
        \centering
        \includegraphics[width=\textwidth]{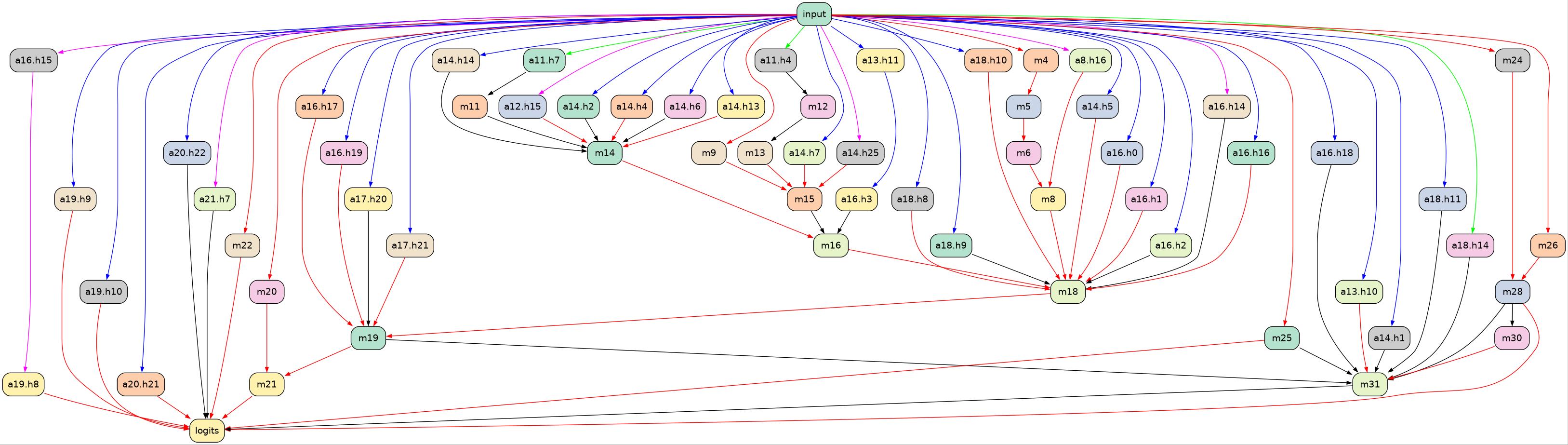}
        \caption{$\mathcal{C}_{\text{Static}}(\theta)$}
        \label{fig:circuit_static}
    \end{subfigure}
    
    \vspace{0.3cm} 
    
    \begin{subfigure}{\textwidth}
        \centering
        \includegraphics[width=\textwidth]{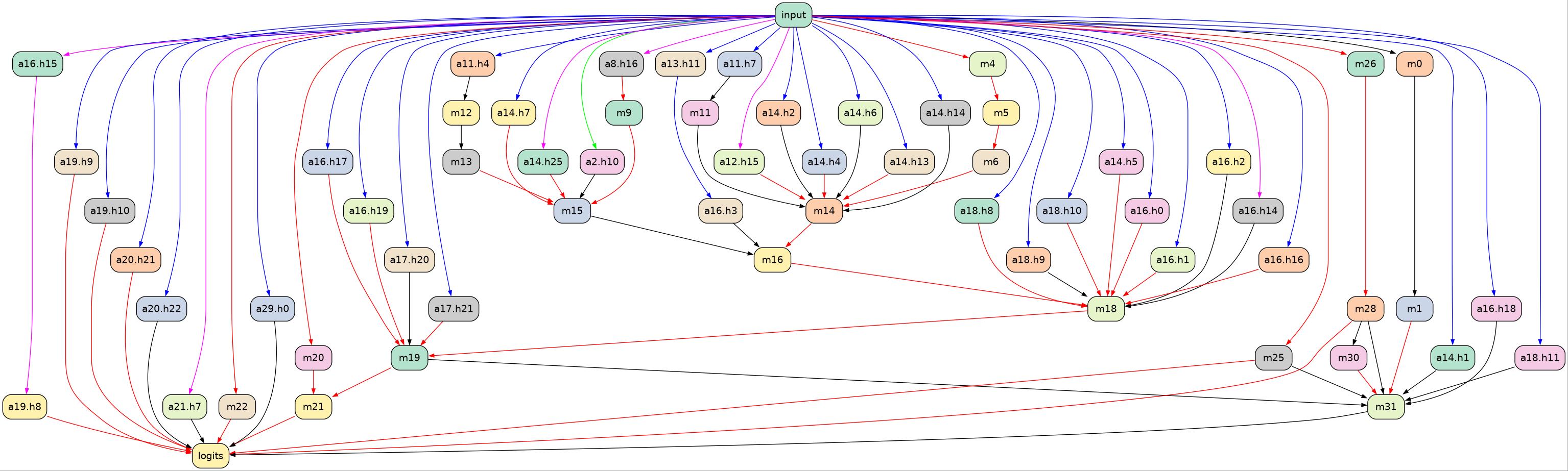}
        \caption{$\mathcal{C}_{\text{Probing}}(\theta')$}
        \label{fig:circuit_probing}
    \end{subfigure}
    
    \vspace{0.3cm} 
    
    \begin{subfigure}{\textwidth}
        \centering
        \includegraphics[width=\textwidth]{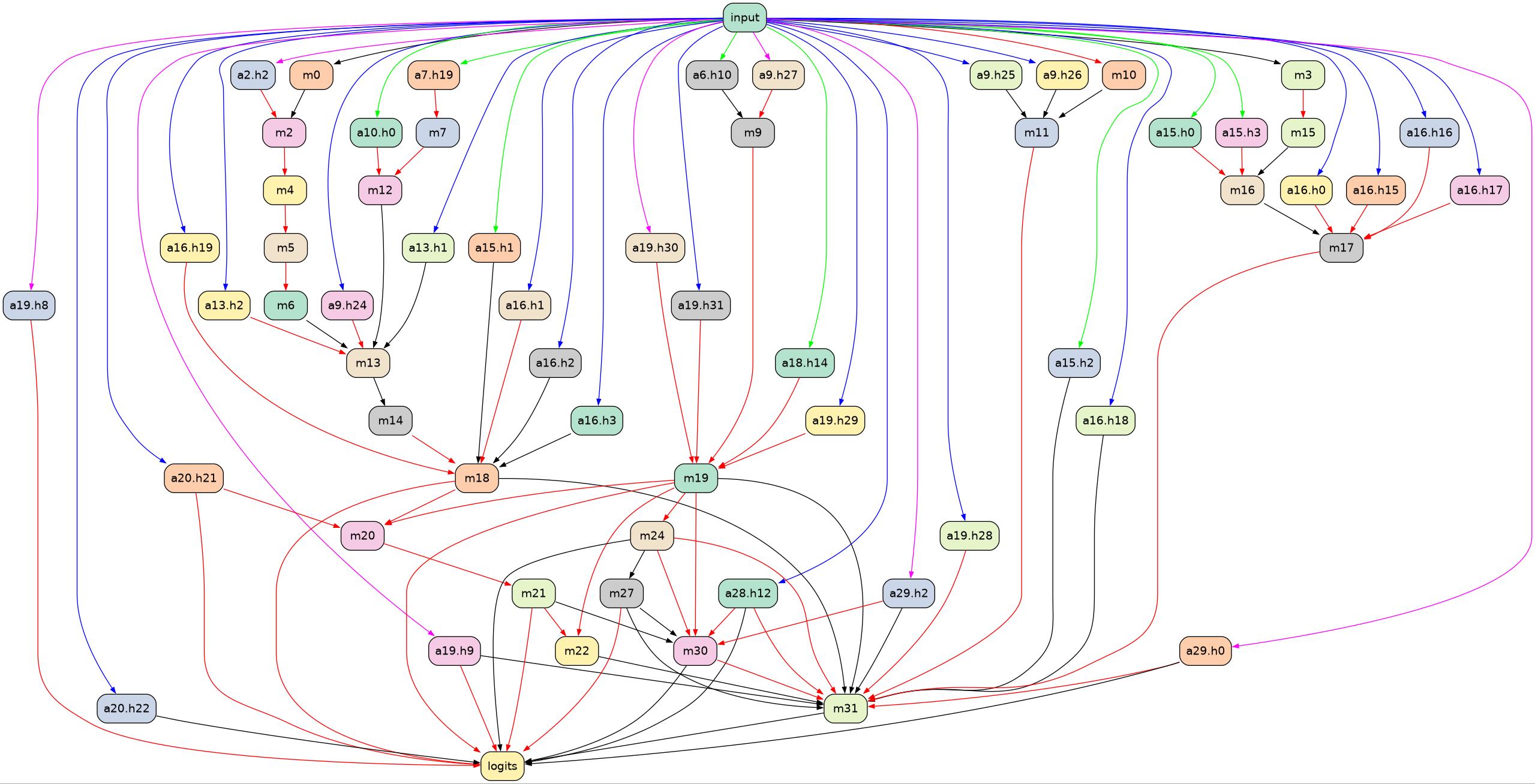}
        \caption{$\mathcal{C}_{\text{Ours}}(\hat{\theta}^{I})$}
        \label{fig:circuit_ours}
    \end{subfigure}
    
    \caption{Circuit graphs corresponding to the three types of parameters.}
    \label{fig:circuit_graphs_all}
\end{figure*}

We provide visualizations of the circuit graphs generated by different localization methods for the IOI task using the Mistral-7B model. As illustrated in Figure~\ref{fig:circuit_graphs_all}, the circuit predicted by our method ($\hat{\theta}^I$) incorporates notably more intermediate nodes compared to those derived from $\theta$ and $\theta'$, leading to more intricate computational pathways from input to output. This structural complexity indicates that our estimated circuit anticipates richer task-specific mechanisms. Notably, these emergent mechanisms are fundamentally absent in the current model parameters, as the base model has not yet fully acquired the target capability.

\section{Dynamic of Learning Factors for Probing SFT}\label{supplearningfactor}
\begin{figure*}[htbp]
    \centering
    \includegraphics[width=0.96\textwidth]{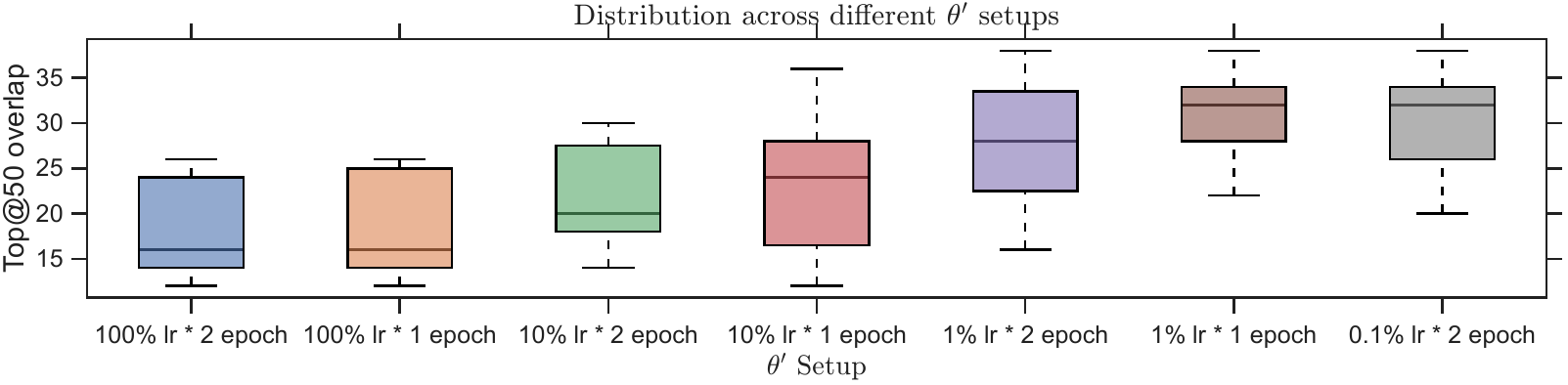}
    \caption{Top 50 overlap with Full-Parameter fine-tuning model across different probing SFT setups. }
    \label{fig:differentprobingsetup2}
\end{figure*}
To empirically justify our hyperparameter selection for the probing SFT, we conduct an additional ablation study on the learning rate and training epochs. We utilize the fully fine-tuned Mistral-7B model as the surrogate ideal model and evaluate the component-level estimation pipeline on the IOI task, fixing the training data size to $1\%$ of the dataset. As illustrated in Figure~\ref{fig:differentprobingsetup2}, we observe a phenomenon that mirrors the findings in Figure~\ref{fig:differentprobingsetup}: reducing the learning rate and the number of epochs—which effectively restricts the magnitude of the parameter update—counterintuitively yields a higher circuit overlap with the ideal model. This observation further corroborates our theoretical assertion in Section~\ref{secprobingsft}: maintaining a sufficiently small parameter update during probing is imperative to preserve the necessary degrees of freedom for the extrapolation term $K \cdot \Delta \theta$. Based on these empirical validations, we finalize our probing SFT configuration as utilizing $1\%$ of the training samples, trained for a single epoch at $1\%$ of the original learning rate.

\end{document}